\documentclass[times,preprint,10pt]{elsarticle}
\journal{Pattern Recognition}
\usepackage[letterpaper,top=4.3cm,right=4.8cm,bottom=4.3cm,left=4.8cm]{geometry}
\usepackage{amssymb}
\usepackage{textcomp}
\usepackage{stfloats}
\usepackage{url}
\usepackage{verbatim}
\usepackage{graphicx}
\usepackage{color, soul}
\usepackage{bbding}
\usepackage[table]{xcolor}
\usepackage{multirow}
\usepackage{amsmath,amsfonts}
\usepackage{algorithmic}
\usepackage{algorithm}
\usepackage{array}
\usepackage{setspace}
\AtBeginDocument{\doublespacing}
\usepackage{booktabs}
\usepackage[caption=false,font=footnotesize,labelfont=rm,textfont=rm]{subfig}
\usepackage{hyperref}
\usepackage{colortbl}
\definecolor{Gray}{gray}{0.93}
\begin{document}
	
	\begin{frontmatter}
		\title{A Spatial-Spectral-Frequency Interactive Network for Multimodal Remote Sensing Classification}
		
		\author[label1]{Hao Liu}
		\author[label2,label3]{Yunhao Gao}
		\author[label2,label3]{Wei Li}
		\author[label4,label5]{Mingyang Zhang}
		\author[label4,label5,label6]{Maoguo Gong}
		\author[label1]{Lorenzo~Bruzzone\corref{cor1}}
		
		\address[label1]{Department of Information Engineering and Computer Science, University of Trento, 38123 Trento, Italy}
		\address[label2]{School of Information and Electronics, Beijing Institute of Technology, Beijing 100081, China}
		\address[label3]{Beijing Key Laboratory of Fractional Signals and Systems, Beijing Institute of Technology, Beijing 100081, China}
		\address[label4]{School of Electronic Engineering, Xidian University, Xi’an 710071, China}
		\address[label5]{Key Laboratory of Collaborative Intelligent Systems of Ministry of Education, Xidian University, Xi’an 710071, China}
		\address[label6]{Academy of Artificial Intelligence, Inner Mongolia Normal University, Hohhot 010028, China }
		\cortext[cor1]{Corresponding author: Lorenzo Bruzzone (\url{lorenzo.bruzzone@unitn.it})}
		
		\begin{abstract}		
			 Deep learning-based methods have achieved significant success in remote sensing Earth observation data analysis. Numerous feature fusion techniques address multimodal remote sensing image classification by integrating global and local features. However, these techniques often struggle to extract structural and detail features from heterogeneous and redundant multimodal images, particularly in label-scarce scenarios. With the goal of introducing frequency domain learning to model key and sparse detail features, this paper introduces the spatial-spectral-frequency interaction network (S$^2$Fin), which integrates pairwise fusion modules across the spatial, spectral, and frequency domains. Specifically, we propose a high-frequency sparse enhancement transformer to refine spectral signatures by adaptively enhancing discriminative high-frequency components. For spatial-frequency interaction, we present a depth-wise strategy: the adaptive frequency channel module fuses low-frequency structural information with enhanced details in shallow layers, while the high-frequency resonance mask amplifies modality-consistent regions in deep layers using phase similarity. In addition, a spatial-spectral attention fusion module bridges the gap between spectral and spatial branches at intermediate depths. Extensive experiments on four benchmark datasets demonstrate that S$^2$Fin exhibits good robustness and generalization, and its performance significantly outperforms state-of-the-art methods in few-sample settings. The code is available at \url{https://github.com/HaoLiu-XDU/SSFin}.
		\end{abstract}
		
	\end{frontmatter}
	
	\section{Introduction}
	\label{sec1}
	Classification of remote sensing imagery enables extraction of Earth-surface information for applications such as environmental monitoring \cite{he2017environmental}, urban planning \cite{ye2025lightweight}, and natural-resource management \cite{gao2022fusion}. Widely used data sources include hyperspectral and multispectral images (HSIs/MSIs) represented by spatial–spectral data cubes \cite{qingyun2022cross}, synthetic aperture radar (SAR) data with all-weather imaging and characterized by the presence of speckle noise \cite{singh2016analysis,singh2021review}, and light detection and ranging (LiDAR) providing high-resolution elevation data \cite{9716784,9999457}. Fusing spectral and active sensor data exploits their complementary strengths to improve classification accuracy and robustness in remote sensing applications \cite{liu2025three,wang2025lmfnet}. 

	Recently, deep learning-based methods have emerged as promising tools for passive and active sensor data classification \cite{9174822}. Methods can be broadly divided into spatial-only and joint spatial–spectral approaches. Rich spatial information from multimodal data prompts spatial fusion-based research, including reconstruction‐based methods \cite{9179756,9598903}, adversarial training strategies \cite{10066307}, representation enhancement approaches \cite{10167673}, and self-supervised learning techniques \cite{xue2025multimodal}. However, limited exploitation of spectral information often degrades classification accuracy. Thus, many studies have focused on spatial-spectral fusion techniques, including spectral sequence transformers \cite{9755059}, masked autoencoders \cite{10314566}, and global-local fusion networks \cite{li2023mixing,tu2024ncglf2}. These methods have achieved promising performance in multimodal classification. However, most existing fusion methods operate purely in the image domain, where structural information and high-frequency details are entangled, often leading to blurred boundaries and degraded feature consistency \cite{10648934}, especially for real-world few-sample remote sensing scenarios \cite{liu2024hybrid}.
	
\begin{figure}[!t]
	\centering
	\includegraphics[width=\textwidth]{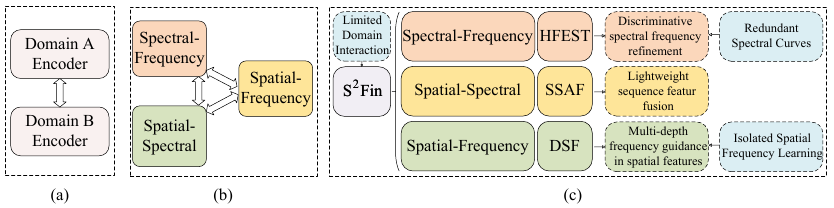}
	\caption{Workflow comparisons and mapping of module motivations. (a) The traditional methods focus on dual-domain fusion from the spatial, spectral, and frequency domains. (b) The proposed spatial-spectral-frequency interaction framework aims to simultaneously perform pairwise interactions between the three domains. (c) The core principles of different modules and their mapping to research gaps.}
	\label{intro1}
	\end{figure}

	In scenarios with limited labeled data, learning robust representations is challenging due to the risk of overfitting redundant spatial features. Spatial-frequency domain techniques address this issue by producing sparse representations that emphasize informative high-frequency components \cite{chen2026f2net}. These components capture critical details such as edges and textures \cite{4135672,qiao2018joint}, which are essential for distinguishing visually similar categories. By focusing on these discriminative features while suppressing redundant information, frequency-domain representations improve sample efficiency and enhance feature extraction \cite{yang2025ffr}. Moreover, spatial-frequency methods enhance spatial modeling from a global perspective \cite{song2025efficient}, making spatial–frequency fusion effective for improving image processing tasks \cite{yu2022frequency}. In the field of remote sensing, research has focused on Fourier transform–based methods \cite{8737724,zhao2022multisource2,10945964}, fractional fusion techniques \cite{zhao2022multisource,9829269} and Gabor filter–based feature extraction approaches \cite{9383794}, wavelet-driven fusion network \cite{11386975}. While multimodal fusion methods have advanced considerably, three important research gaps remain:
	
	\begin{enumerate}
		\item Limited domain interaction: As illustrated in Fig. \ref{intro1}(a), conventional methods prioritize dual-domain interactions (e.g., spatial-spectral or spatial-frequency). Local interactions limit the ability to jointly exploit the global structure of multimodal images and discriminative details \cite{10648934}.
		\item Redundant spectral curves: High similarity and continuity among spectral bands in HSIs result in difficulty for extracting optimal features \cite{qiao2018joint}. Existing methods overemphasize various attention mechanisms and network structures while neglecting the decomposition of spectral signals from the frequency domain, which can elegantly capture subtle inter-class differences in spectral data.
		\item Isolated spatial frequency Learning: High- and low-level network features correspond to object-level semantics and fine-grained background textures, respectively. Correspondingly, low-frequency components effectively encode global structure and semantics, whereas high-frequency components capture fine details \cite{sun2024high}. However, existing fusion strategies often apply a "one-size-fits-all" approach, such as preprocessing transformers, which ignores frequency guidance at multiple network depths.
	
	\end{enumerate}

	Motivated by these challenges, we propose the spatial-spectral-frequency interactive network (S$^2$Fin) that improves pixel-level, few-sample multimodal remote sensing classification. As shown in Fig. \ref{intro1}(b)(c), unlike previous work on dual-domain fusion, S$^2$Fin aims to enhance the frequency interactions along both the spectral and spatial dimensions at multiple network depths. S$^2$Fin comprises three components: a high-frequency enhancement and sparse transformer (HFEST) for spectral–frequency interaction, a spatial–spectral attention fusion module (SSAF), and a depth-wise spatial frequency fusion strategy (DSF). HFEST enhances informative high-frequency spectral components by learning adaptive frequency filters to mitigate the contribution of redundant spectral curves. DSF further performs spatial–frequency learning through an adaptive frequency channel module (AFCM) and a high-frequency resonance mask (HFRM), where AFCM enhances high-frequency details while preserving shared low-frequency structures in shallow layers, and HFRM further strengthens representations at key spatial locations in deeper layers. Fig.~\ref{intro1}(c) illustrates the motivations of these modules. To simulate the few-sample scenario, we operate in a supervised training set where 10 labeled samples are randomly selected for each class. The main contributions can be highlighted as follows.

	\begin{enumerate}
		\item We propose the S$^2$Fin, a novel multimodal remote sensing classification framework, integrating pairwise fusion and frequency enhancement modules across spatial, spectral, and frequency domains.
		
		\item We introduce the HFEST to extract key spectral features from frequency domain. This module employs a sparse attention mechanism to improve the estimation of high-frequency filter's parameters, thereby enabling discriminative spectral frequency refinement.
		
		\item We present a depth-wise spatial frequency fusion strategy utilizing the AFCM and HFRM in shallow and deep network layers, respectively. The AFCM fuses low-frequency structural information and enhances high-frequency modality-specific details by balancing channel attention. The HFRM amplifies specific amplitude regions based on phase similarity, strengthening the focus on modality-common areas.
		
	\end{enumerate}
	
	In summary, the primary objective is to establish a novel S$^2$Fin framework that enables the classification of spectral images and SAR/LiDAR multimodal remote sensing data under limited samples through synergistic spatial, spectral, and frequency interactions. This unified design alleviates heterogeneous feature extraction and labeled data scarcity, enabling robust and efficient classification across diverse sensor pairs for complex Earth observation tasks.
	
	The remainder of this paper is organized as follows. Section \ref{Background and Motivation} provides background knowledge about S$^2$Fin. Section \ref{methodology} describes the proposed method. Section \ref{experiment} validates the effectiveness of S$^2$Fin on four remote sensing datasets and analyzes the related hyperparameters. Finally, Section \ref{conclusion} draws the conclusions of this paper.
	
\section{Related Work}
\label{Background and Motivation}
This section first reviews the background and advanced methods of frequency domain learning, then introduces related techniques of multimodal feature fusion.

\subsection{High-Frequency Enhancement}

Frequency domain transformations are widely used methodology for converting signals from their original temporal or spatial representations into a form that expresses frequency components \cite{sun2024high,1456290}. Frequency domain transform can analyze the amplitude, phase, and frequency distribution of a signal to achieve the various tasks including filtering, noise reduction, and feature extraction \cite{yu2022frequency,xu2020learning}. 

In the spatial frequency domain of an image, low-frequency components typically correspond to the smooth areas, whereas high-frequency components correspond to the rapidly changing parts, such as edges, textures, and details \cite{4135672,10943595}. In the literature, several techniques focused on the high-frequency enhancement to extract key features. Sun et al. \cite{sun2024unsupervised} utilized an high-frequency enhancement module to capture details present in the images. Behjati et al. \cite{9778017} proposed a frequency-based enhancement block to enhance the part of high frequencies while forward the low-frequencies. Wang et al. \cite{wang2023learning} employed fast Flourier convolution with attention mechanism in the high-frequency domain. In addition, some studies have attempted to add adaptive thresholds to smoothing filters \cite{singh2018new}, correlation fusion \cite{singh2021new}, and wavelet transform \cite{singh2020new} for feature processing of remote sensing images.

In the frequency domain, phase information describes the position and structure of the various frequency components within an image. It encodes the relative positions of different frequency components, serving as a key carrier of image structural information \cite{1456290}. This work aims to utilize high-frequency enhancement methods and phase information to build spatial mask for multimodal feature extraction.

\subsection{Multimodal Image Classification}
Multimodal learning integrates complementary information from different data sources, resulting in robust and reliable outcomes in various tasks. In remote sensing data classification, deep learning multimodal architectures, primarily based on CNNs and Transformers, are increasingly popular.

CNNs effectively capture local features and are widely used for multimodal data fusion \cite{hong2021multimodal}. For example, Wu et al. \cite{9598903} introduced a CNN backbone with a cross-channel reconstruction module, while Gao et al. \cite{10066307} proposed an adversarial complementary learning strategy within a CNN model. Wang et al. \cite{10167673} developed a representation-enhanced status replay network. However, although these techniques excel at detecting local features, their strong local sensitivity and lack of long-range dependency modeling limit their ability to capture rich contextual information.

Due to its powerful global perception, the Transformer has recently been applied to the fusion of multimodal remote sensing imagery. For instance, Xue et al. \cite{9755059} proposed a deep hierarchical vision Transformer, and Zhou et al. \cite{zhou2024tcpsnet} employed a four-branch deep feature extraction framework with a dynamic multi-scale feature extraction module for multimodal joint classification, while Ni et al. \cite{10438852} introduced a multiscale head selection Transformer.

Recently, Mamba has attracted attention for multimodal fusion because of its efficient training and inference capabilities \cite{xie2024fusionmamba}. In the field of remote sensing, there are studies on spatial-spectral Mamba \cite{10738515} and multi-scale Mamba \cite{10856240}. The Mamba architecture uses the state space model to capture long-range dependencies, which reduces computational requirements and is suitable for long sequence tasks \cite{yu2024mambaout}. Meanwhile, the transformer focuses on global features based on the attention mechanism.
This work fuse multimodal data based on Mamba and transformer techniques to achieve long-range dependency feature fusion and save computing resources.

\section{Methodology}
\label{methodology}
\begin{figure}[!t]
	\centering
	\includegraphics[width=\textwidth]{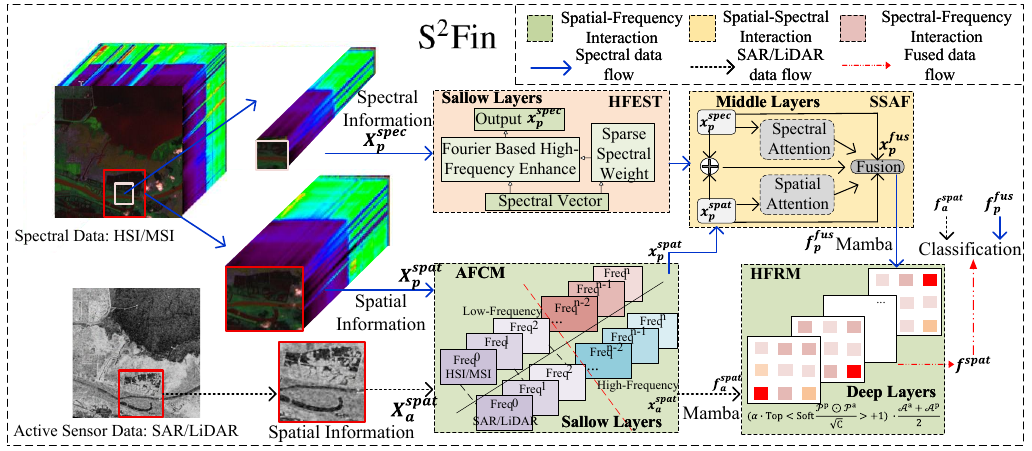}
	\caption
	{Illustration of the proposed S$^2$Fin framework.}
	\label{Framwork}
\end{figure}

\subsection{Overall Framework of S$^2$Fin}
The S$^2$Fin framework follows a hierarchical interaction pipeline that progressively fuses spatial, spectral, and frequency information across different depths of the backbone. As illustrated in Fig. \ref{Framwork} (see Supplementary Material A for a detailed overview), the process begins in the shallow layers, where the spectral branch utilizes HFEST to enhance sparse high-frequency details, while the spatial branches employ AFCM to share global low-frequency structures across modalities and preserve distinctive textures. In the middle layers, SSAF cross-modulates attention between spectral and spatial branches to enable spatial–spectral exchange. Finally, in the deep layers, HFRM uses phase resonance to produce a high-frequency mask that filters noise and highlights consistent semantic structures for classification.

Let $X$, $x$, and $f$ represent data features at different depths, $spec$ and $spat$ represent spectral and spatial data, and $p$ and $a$ represent passive and active images, namely spectral data and SAR/LiDAR, respectively.

For clarity, “frequency” here means transform-domain cues used along two axes. (1) Spectral frequency refers to frequency components obtained by transforming the spectral signal along the spectral dimension of hyperspectral or multispectral data, which highlights variations across spectral bands. (2) Spatial frequency refers to frequency components derived from spatial feature maps through 2D transforms, where low-frequency components encode global structure while high-frequency components capture edges and textures. The spatial-frequency representation can be decomposed into amplitude, which describes the strength of a frequency component, and phase, which encodes structural alignment and spatial location.

In the next subsections, the modules included in the S$^2$Fin framework are described in detail, offering insights into their functionalities.

	\begin{figure}[t]
	\centering
	\includegraphics[width=0.9\linewidth]{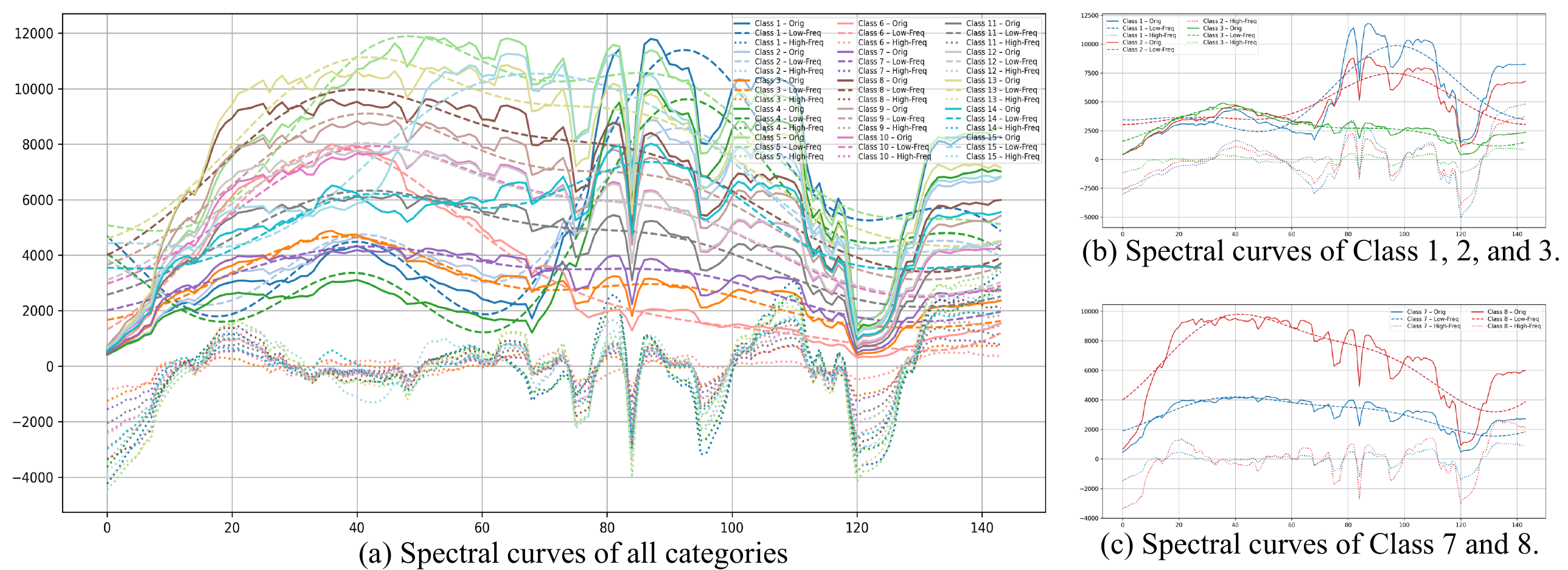}
	\caption
	{Spectral curves filtered by low- and high-frequency components of the HSI of the Houston dataset obtained via 1D discrete Fourier transform. The horizontal axis represents the number of bands and the vertical axis represents the reflectivity values. (a) All categories. (b) Categories 1 (healthy grass), 2 (stressed grass), and 3 (synthetic grass). (c) Categories 7 (residential) and 8 (commercial). Note that low-frequency curves show substantial overlap across classes, while high-frequency curves exhibit clearer inter-class separation, visually demonstrating that high-frequency components carry more discriminative local spectral detail than low-frequency components.}
	\label{spectral}
\end{figure}
	
	\subsection{Spectral-Frequency Modeling: High-Frequency Enhancement and Sparse Transformer}

	Remote sensing objects exhibit spectral signatures that are both complex and closely similar, making it challenging to characterize their spectral-dimensional features. Frequency-domain analysis decomposes a spectral signal into low-frequency components, which are smooth and highly correlated, and high-frequency components that exhibit larger variations. As illustrated in Fig. \ref{spectral}, we analyze category-distinguishing information by applying a one-dimensional discrete Fourier transform (DFT) along the spectral axis and reconstructing high- and low-frequency filtered versions of each spectral feature. Low-frequency components mainly encode global spectral structure shared across many materials, causing different classes to exhibit similar low-frequency patterns. High-frequency components instead capture rapid spectral variations caused by material boundaries and fine textures. These variations tend to increase inter-class differences while remaining relatively consistent within each class, making them more discriminative when labeled samples are limited. Consequently, emphasizing high-frequency information helps the model separate classes more effectively under scarce supervision. Figs. \ref{spectral}(b)(c) compare specific categories, highlighting this disparity more clearly.

	\begin{figure}[t]
		\centering
		\includegraphics[width=0.8\textwidth]{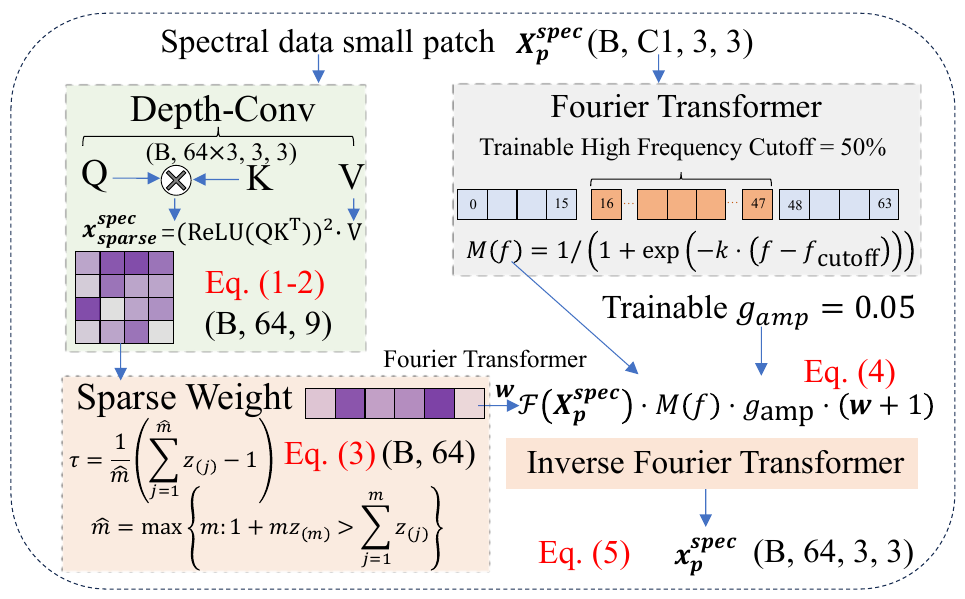}
		\caption
		{Structure of HFEST. The left part represents the high-frequency enhancement branch, while the right part is the sparse attention branch. The two branches are merged through a linear layer and a norm layer.}
		\label{HFES}
	\end{figure}
	
	Motivated by these observations, the HFEST mainly utilizes a sparse spatial-spectral attention to enhance the high-frequency filter's parameters, as shown in the Fig. \ref{HFES}. Initially, HSI and MSI have multiple spectral channels, which especially in HSI may have high similarity and redundancy. We combine depth-wise convolution and squared ReLU-based attention to remove the similarities with negative relevance from the spectral dimension. 
	
	First, we obtain the $\boldsymbol{Q}$, $\boldsymbol{K}$, and $\boldsymbol{V}$ required for attention through depth-wise convolution, which captures spectral relationships within individual channels:
	\begin{equation}
		\begin{aligned}
			\label{1-1}
			\boldsymbol{Q}, \boldsymbol{K}, \boldsymbol{V} = {split} (\text{Depth-Conv}(\boldsymbol{X^{spec}_p})),
		\end{aligned}
	\end{equation}
	where ${split}$ divides the depth-wise convolution tensor into attention vectors. The spectral features $\boldsymbol{x^{spec}_{sparse}}$ after sparse attention processing can be expressed as:
	\begin{equation}
		\begin{aligned}
			\label{1-2}
			\boldsymbol{x^{spec}_{sparse}} = \text{ReLU}^2(\boldsymbol{Q} \boldsymbol{K}^\top) \cdot \boldsymbol{V},
		\end{aligned}
	\end{equation}
	where $\text{ReLU}^2$ represents squared ReLU activation function. By applying a sparse method, the model focuses on the informative spectral features instead of redundant hyperspectral bands.
	
	To achieve spatial sparsity, we employ a differentiable projection. For a sorted coefficient vector $\boldsymbol{z}_{(1)} \ge \dots \ge \boldsymbol{z}_{(M)}$, we identify the support size $\hat{m}$ and the adaptive threshold $\tau$ as:
	\begin{equation}
		\hat{m} = \max \left\{ m : 1 + m z_{(m)} > \sum_{j=1}^m z_{(j)} \right\}, \quad \tau = \frac{1}{\hat{m}} \left( \sum_{j=1}^{\hat{m}} z_{(j)} - 1 \right)
	\end{equation}
	
	The final sparse weight $w_i$ is obtained by a ReLU-like truncation $w_i = \max(0, z_i - \tau)$. Note that this projection is piecewise linear and ensures end-to-end differentiability, as the gradient flows through the support set $\hat{m}$ via the threshold $\tau$, similar to the sub-gradient properties of the ReLU activation.
	
	Furthermore, to overcome the gradient breakage problem caused by traditional hard truncation, we introduce a differentiable soft mask $M(f)$ based on the Sigmoid function. This mask defines the frequency weights through a trainable cutoff parameter $f_{\text{cutoff}}$:
	 \begin{equation}
	 M(f) = 1/\left(1 + \exp(-k \cdot (f - f_{\text{cutoff}}))\right), 
	 \end{equation}
	 where $k$ is a large scaling factor and $f \in [0, 1]$ represents the components from low to high frequency after normalization. In this case, the low-frequency components are very close to zero, but the process is still differentiable, thus achieving approximate low-frequency suppression. The trainable thresholds $f_{\text{cutoff}}$ and gain coefficients $g_{\text{amp}}$ are added to the transform, and the values are automatically updated as the network iterates. This process can be expressed as:
	\begin{equation}
		\begin{aligned}
			\label{1-5}
			\mathcal{F}'(\boldsymbol{X^{spec}_p}) = \mathcal{F}(\boldsymbol{X^{spec}_p})\cdot M(f) \cdot g_{\text{amp}} \cdot (\boldsymbol{w}+1),
		\end{aligned}
	\end{equation}
	where $\mathcal{F}$ and $f$ are the Fourier transform and frequency component, respectively. Fourier transform is used because it naturally decomposes the spectral signature and allows straightforward frequency separation. After inverse Fourier transform ${\mathcal{F}'}^{-1}$, we can get the enhanced high-frequency components $\boldsymbol{x^{spec}_{hf}}$. The output of the HFEST is obtained as:
	\begin{equation}
		\begin{aligned}
			\label{1-6}
			\boldsymbol{x^{spec}_{p}} = FC \cdot(\boldsymbol{x^{spec}_{sparse}}, \boldsymbol{x^{spec}_{hf}}))+\boldsymbol{X^{spec}_p},
		\end{aligned}
	\end{equation}
	where $FC$ represents a linear layer.

	\subsection{Spatial-Frequency Modeling: Depth-Wise Spatial Frequency Fusion Strategy}
	
\begin{figure}[tbp]
		\centering
		\includegraphics[width=0.85\linewidth]{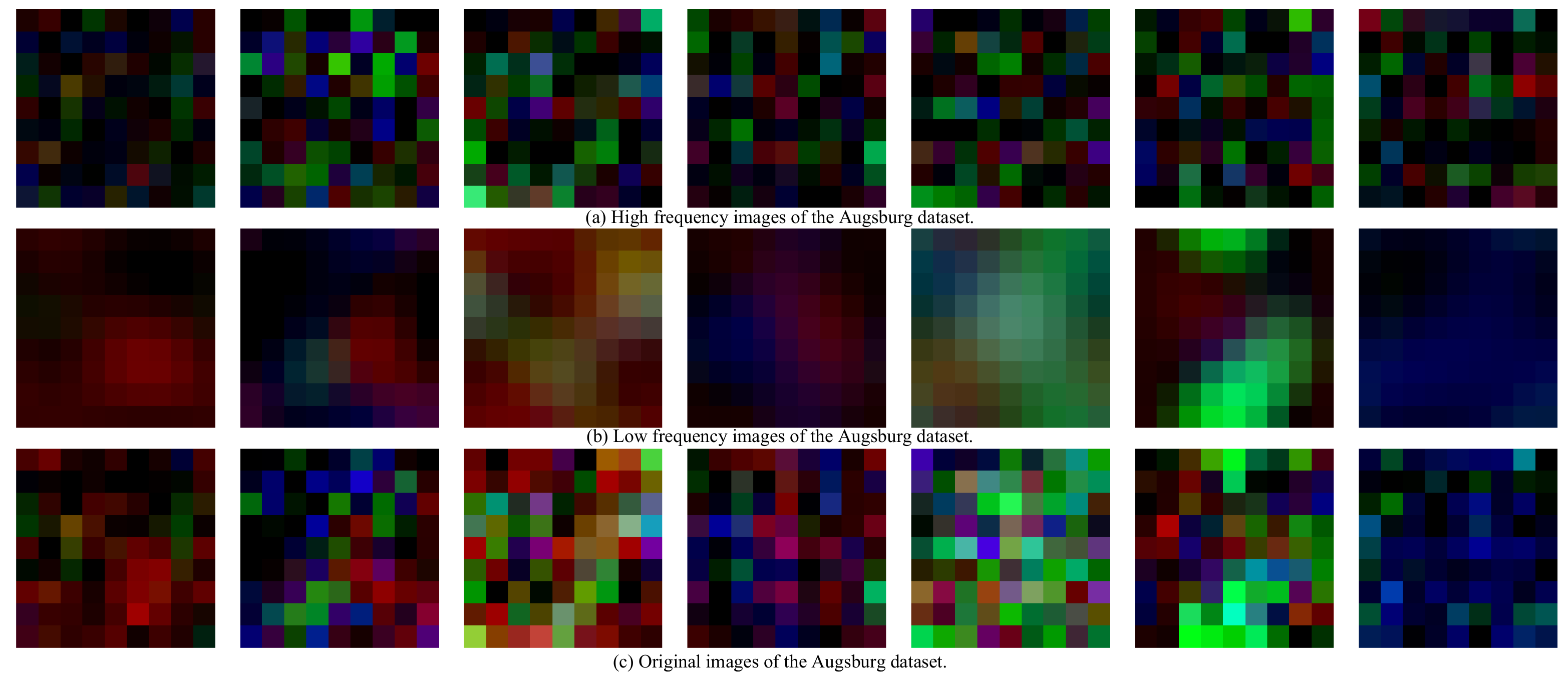}
		\caption
		{Example of images of the HSIs of the Augsburg dataset filtered by low- and high-frequency componentsobtained by applying a 2D DFT along the spatial dimension. Three main bands are selected following principal component analysis (PCA), and ten samples per class are processed by DFT to generate average component magnitude images. The seven class-averaged images are displayed from left to right.}
		\label{spatial}
	\end{figure}

	The two-level spatial-frequency fusion strategy is designed to separately extract semantic category information and boundary details from different network layers \cite{10648934}. As illustrated in Fig. \ref{spatial}, low-frequency components typically capture the structural information of ground objects, whereas high-frequency components encode fine-grained category-specific details. This strategy incorporates the AFCM for low-level channel attention and the HFRM for high-level spatial amplitude resonance.
	
	\subsubsection{Shallow Layers: Adaptive Frequency Channel Module}

	A fundamental step in our methodology is the transformation of spatial features into the frequency domain to enable feature recalibration based on frequency content. In the shallow stages, we implement the regularized 2D discrete cosine transform (DCT) for channel-dimension operations as it is real-valued and provides strong energy compaction for local channel-wise structure. Given a single-channel input $\boldsymbol{x} \in \mathbb{R}^{H \times W}$, it can be defined as:
	\begin{equation}
		\begin{aligned}
			\boldsymbol{f_{h,w}^{Fre}} = C(h)\cdot C(w) \sum_{i=0}^{H-1} \sum_{j=0}^{W-1} \boldsymbol{x_{i,j}} \cos\left(\frac{\pi h}{H}\left(i+\frac{1}{2}\right)\right)\cos\left(\frac{\pi w}{W}\left(j+\frac{1}{2}\right)\right),\\
			s.t. \;\; h \in \{0,1,\cdots,H-1\}, w \in \{0,1,\cdots,W-1\},
			\label{eq:dct_formal}
		\end{aligned}
	\end{equation}
	where $\boldsymbol{f^{Fre}} \in \mathbb{R}^{H \times W}$ is the resulting frequency spectrum. The normalization coefficients $C(u)$ are given by $C(u) = \sqrt{1/N}$ for $u=0$ and $C(u) = \sqrt{2/N}$ for $u>0$, where $N$ represents the length of the dimension. This ensures the orthogonality of the transform.
	
	The central motivation of the AFCM is that the frequency spectrum can be partitioned to disentangle shared, structural information from modality-specific details. Low-frequency coefficients encode global structure and are amenable to joint cross-modal processing, whereas high-frequency coefficients capture fine texture and should be preserved modality-specifically to retain unique characteristics.
	
	This principle is mathematically realized as follows. Given two multimodal spatial feature maps, $\boldsymbol{X^{spat}_p}$ and $\boldsymbol{X^{spat}_a}$, corresponding to spectral and active sensor data, the modulated output $\boldsymbol{x^{spat}_{p}}$ for the passive modality is computed by:
	\begin{equation}
		\begin{split}
			\boldsymbol{x^{spat}_{p}} &= \boldsymbol{X^{spat}_p} \odot \Bigl(
			\sigma \cdot FC \:(\mathcal{P}_{\text{high}}\:(\operatorname{DCT}\:(\boldsymbol{X^{spat}_p}))) \\
			&\: + \sigma \cdot FC\:\Bigl(\frac{\mathcal{P}_{\text{low}}\:(\operatorname{DCT}\:(\boldsymbol{X^{spat}_p})) +
				\mathcal{P}_{\text{low}}\:(\operatorname{DCT}\:(\boldsymbol{X^{spat}_a}))}{2}\Bigr) + 1 \Bigr),
		\end{split}
		\label{eq:afcm_formal}
	\end{equation}
	where $\odot$ denotes the element-wise Hadamard product. The operators $\mathcal{P}_{\text{low}}(\cdot)$ and $\mathcal{P}_{\text{high}}(\cdot)$ represent the frequency partitioning functions, which extract vectors of low- and high-frequency coefficients from a given spectrum based on predefined index sets. $\sigma$ is the sigmoid activation function. The first term of this formula is used to enhance the high-frequency term of the input, while the second term represents the low-frequency term fused with other source. The corresponding output for the active modality $\boldsymbol{x^{spat}_{a}}$ is obtained through a symmetrical application of Eq.~\eqref{eq:afcm_formal}. This mechanism thereby allows the network to dynamically fuse shared structural knowledge while concurrently enhancing distinguishing modality-specific information.

	\subsubsection{Deep Layers: High-Frequency Resonance Mask}
	
	On the one hand, to amplify the common information of multimodal images, we try to find the high-frequency regions of each modality as shown in Fig \ref{spatial}(a), and enhance these similar regions. On the other hand, the semantic information in the deep layers of the network is highly correlated with the classes to be recognized. The HFRM is designed to amplify the detail features. We use the simple and flexible 2D Fourier transform to decompose the spatial features $\boldsymbol{f^{fus}_p}$ and $\boldsymbol{f_{a}^{spat}}$ to obtain the amplitude and phase:
	\begin{equation}
		\begin{aligned}
			\boldsymbol{\mathcal{A}^{p}}, \boldsymbol{\mathcal{P}^{p}}= \mathcal{F}\boldsymbol{(f^{fus}_p)}, \quad
			\boldsymbol{\mathcal{A}^{a}}, \boldsymbol{\mathcal{P}^{a}} = \mathcal{F}\boldsymbol{(f_{a}^{spat})}.
			\label{3-2-3}
		\end{aligned}
	\end{equation}
	
	The amplitude represents the intensity of the various frequency components within an image. Enhancing the amplitude in the high-frequency areas improves the image's details and the edge features \cite{4135672}. Intuitively, the HFGM locates the significant high-frequency parts within an image by leveraging the phase correlations of multimodal data, and subsequently enhances the image detail information by amplifying the amplitude. 
	
	To simulate the coherent resonance effect of multimodal features in local space, we designed a differentiable $Top<Soft (\cdot, T)>$ selection operator based on the Softmax function. It has an extremely low temperature $T$=0.01 so that the network can automatically locate the spatial frequency points with the highest phase correlation in an end-to-end manner and enhance their amplitudes. This design retains the physical intuition of hard attention while ensuring the stability of model optimization through gradient flow. Given the correlation scores $z_i$ between multimodal features, the operator is defined as:
	\begin{equation}
		Top<Soft(z_i, T)> = \frac{\exp(z_i / T)}{\sum_{j=1}^n \exp(z_j / T)}.
	\end{equation}
	
	When $T \to 0$, the distribution approaches a one-hot vector, performing a "Top-1" selection of the strongest resonance point. Based on this, the amplitudes with high attention value are intensified:
	\begin{equation}
		\begin{aligned}
			\label{3-2-12}
			\boldsymbol{\mathcal{A}}=(\alpha \cdot Top<Soft \frac{\boldsymbol{\mathcal{P}^{p}} \odot \boldsymbol{\mathcal{P}^{a}}}{\sqrt {C}}>+1) \cdot \frac{\boldsymbol{\mathcal{A}^{p}}+\boldsymbol{\mathcal{A}^{a}}}{2},
		\end{aligned}
	\end{equation}
	where $C$ refers to the number of channels, $\boldsymbol{\mathcal{A}}$ represents the final integrated amplitude, and $\alpha$ is a trade-off parameter.
	
	\begin{figure}[t]
		\centering
		\includegraphics[width=0.9\textwidth]{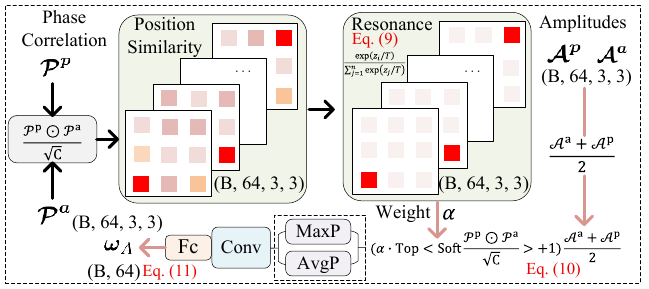}
		\caption
		{Illustration of the generation process of $\boldsymbol{\omega_{A}}$. First the operations with black arrows are applied, then the operations with red arrows are applied.}
		\label{wa}
	\end{figure} 
	
	To further eliminate noise and extract high-level semantic information that is beneficial for classification, further processing of the amplitude is undertaken:
	\begin{equation}
		\begin{aligned}
			\label{3-2-114}
			\boldsymbol{\omega_{A}}=Soft \cdot FC \cdot Conv \cdot (\text{MaxP}(\boldsymbol{\mathcal{A}}), \text{AvgP}(\boldsymbol{\mathcal{A}})),
		\end{aligned}
	\end{equation}
	where $MaxP$ and $AvgP$ denote the operations of maximum and average pooling, respectively, and $Conv$ represents a two-dimensional convolution operation. The perception process of $\boldsymbol{\omega_{A}}$ is depicted in Fig. \ref{wa}, where different colors represent distinct spatial weight values, with the top used to select the positions of the highest values.
	
	Finally, the resulting integrated amplitude and phase can be written as:
	\begin{equation}
		\begin{aligned}
			\label{3-2-14}
			\boldsymbol{\mathcal{A}_{fusion}}&=(1+\boldsymbol{\omega_{A}})\cdot(\boldsymbol{\mathcal{A}^{p}}+\boldsymbol{\mathcal{A}^{a}})/{2}, \quad
			\boldsymbol{\mathcal{P}_{fusion}}&=Conv \cdot (\boldsymbol{\mathcal{P}^{p}}+\boldsymbol{\mathcal{P}^{a}})/{2}.
		\end{aligned}
	\end{equation}
	
	After inverse transform $\mathcal{F}^{-1}$ , we can obtain the multimodal spatial features $\boldsymbol{f^{spat}}$ for classification.

	\subsection{Spatial-Spectral Modeling: Spatial-Spectral Attention Fusion}
	
\begin{figure}[t]
		\centering
		\includegraphics[width=0.7\textwidth]{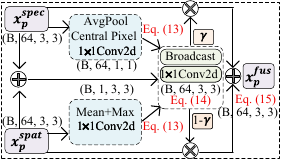}
		\caption
		{Flowchart of the proposed SSAF.}
		\label{HFAF}
	\end{figure}

	SSAF attempts to extend the spectral attention score obtained by HFEST to spatial data, while applies the attention score from AFCM, thereby synthesizing spatial-spectral interaction features. Fig. \ref{HFAF} shows the network structure. 
	
	With $\boldsymbol{x^{spat}_{p}}$ and $\boldsymbol{x^{spec}_{p}}$ in Eq. ~\eqref{eq:afcm_formal} as input, the integrated attention scores are:
	\begin{equation}
		\begin{aligned}
			\boldsymbol{Atte^{spat}}&= \sigma \cdot Conv \cdot( \frac{1}{H \cdot W} \sum_{h=1}^H \sum_{w=1}^W \boldsymbol{x^{spat}_p}, cent (\boldsymbol{x^{spat}_p})),\\
			\boldsymbol{Atte^{spec}} &=\sigma \cdot Conv \cdot(\frac{1}{C} \sum_{c=1}^C \boldsymbol{x^{spec}_{p}}, \max_{c \in \{1, \dots, C\}} \boldsymbol{x^{spec}_{p}}),
		\end{aligned}
	\end{equation}
	where $cent$ represents the spatial center feature and $\max_{c \in \{1, \dots, C\}}$ represents the maximum spectral feature of the channel dimension. Then fused features and attention scores are:
	\begin{equation}
		\begin{aligned}
			\boldsymbol{x^{ss}_p} &= \boldsymbol{x^{spat}_p} + \boldsymbol{x^{spec}_p},\\
			\boldsymbol{Atte^{fus}} &= Broadcast(\boldsymbol{Atte^{spat}}, \boldsymbol{Atte^{spec}}),
		\end{aligned}
	\end{equation}
	where $Broadcast$ expands the attention scores to the entire feature map along the channel and spatial dimensions. Subsequently, the output of SSAF can be written as:
	\begin{equation}
		\begin{aligned}		
			\boldsymbol\gamma &= \sigma \cdot (Conv \cdot(\boldsymbol{x^{ss}_p}, \boldsymbol{Atte^{fus}} ))), \\	
			\boldsymbol{x_p^{fus}} &= \sigma \cdot Conv (\boldsymbol{x^{ss}} + \boldsymbol\gamma \cdot \boldsymbol{x^{spec}_p} + (1 - \boldsymbol\gamma) \cdot \boldsymbol{x^{spat}_p} ). \\
		\end{aligned}
		\label{23}
	\end{equation}
	
	Furthermore, we employ the Mamba module $SSM(\cdot)$ \cite{xie2024fusionmamba} to extract long-range dependency features and refine their fusion via an attention mechanism:
	\begin{equation}
			\label{3-2-17}
			\boldsymbol{f^{fus}_p} = SSM(\boldsymbol{x^{fus}_p}) \cdot \sigma \cdot FC \cdot \left( cent(SSM(\boldsymbol{x^{spat}_p})) + \frac{1}{C} \sum_{c=1}^C SSM(\boldsymbol{x^{fus}_p})\right) , 	
	\end{equation}
	where $\boldsymbol{f^{fus}_p}$ is used for classification and the HFRM. $\boldsymbol{f^{spat}_{p}}$ can be also obtained using $SSM(\boldsymbol{x^{spat}_{p}})$.
	
	Finally, we combine the fused multimodal features $\boldsymbol{f^{spat}}$ from HFRM, the spectral features $\boldsymbol{f^{fus}_p}$ from SSAF, and the SAR/LiDAR features $\boldsymbol{f^{spat}_a}$ from AFCM for classification. Please refer to Supplementary Material A for detailed information.
	
	\section{Experimental Results and Discussion}
	\label{experiment}
	\definecolor{m1}{HTML}{FF0000}
	\definecolor{m2}{HTML}{00FF00}
	\definecolor{m3}{HTML}{0000FF}
	\definecolor{m4}{HTML}{FFFF00}
	\definecolor{m5}{HTML}{00FFFF}
	\definecolor{m6}{HTML}{FF00FF}
	\definecolor{m7}{HTML}{C0C0C0}
	\definecolor{m8}{HTML}{808080}
	\definecolor{m9}{HTML}{800000}
	\definecolor{m10}{HTML}{808000}
	\definecolor{m11}{HTML}{008000}
	\definecolor{m12}{HTML}{800080}
	\definecolor{m13}{HTML}{008080}
	\definecolor{m14}{HTML}{0000FF}
	\definecolor{m15}{HTML}{FFA500}
	
	This section briefly introduces the multimodal remote sensing dataset and experimental setup. Then, it describes the parameter tuning and	ablation study. Next, it presents quantitative and qualitative results, uncertainty and robustness analysis, and cross-region generalization analysis, and discusses the computational complexity.
	
Comparisons have been done against a range of classic and advanced state-of-the-art multimodal remote sensing classification methods. These methods fall into four main groups. (1) Attention-based spectral–spatial fusion: approaches that learn where to attend across spectra and space to improve discrimination (FusAtNet \cite{Mohla_2020_CVPR_Workshops}). (2) Modality-aware architectural fusion: network designs that account for different sensor properties or combine complementary backbones (AsyFFNet \cite{9716784}, Fusion-HCT \cite{9999457}, MACN \cite{li2023mixing}). (3) Learning and alignment strategies: training schemes that align modalities or reinforce robustness via coupled learning and contrastive objectives (CALC \cite{lu2023coupled}, UACL \cite{10540387}). (4) Multi-scale and global–local aggregation: methods that fuse information at multiple scales or explicitly combine global and local features to retain context and fine details (NCGLF \cite{tu2024ncglf2}, MSFMamba \cite{10856240}).

	\subsection{Description of Datasets}

	\begin{table*}[htbp]
		\centering
		\caption{Summary of dataset characteristics and OA improvement (\%)}
		\resizebox{\textwidth}{!}{
			\begin{tabular}{lccccccccc}
			\toprule
			Dataset&Area Description & Modalities & Channels & Spatial Size & Classes &Numbers& Top-Baseline & S$^2$Fin  & $\Delta$\\
			\midrule
			Houston 2013 \cite{hong2021multimodal}&Urban campus, Houston & HSI + LiDAR & 144 + 1 & $349 \times 1905$ & 15& 15029& 87.83 & 89.19 & +1.36 \\
			Augsburg \cite{hong2021multimodal} &Rural landscape, Augsburg & HSI + SAR & 180 + 4 & $332 \times 485$ & 7 &78294& 77.67 & 79.91 & +2.24 \\
			Yellow River Estuary \cite{gao2022fusion} &Wetlands, Shangdong& HSI + SAR & 166 + 4 & $960 \times 1170$ & 5&464671 & 65.34 & 67.54 & +2.20 \\
			LCZ HK \cite{9174822}&Urban and rural areas, Hong Kong&MSI + SAR & 10 + 4 & $529 \times 528$ & 13 &8846& 71.87 & 72.26 & +0.39 \\
			\bottomrule
		\end{tabular}}
		\label{tab:dataset_comparison}
	\end{table*}
	Table \ref{tab:dataset_comparison} provides a comprehensive overview of the four benchmark multimodal datasets utilized in this study, detailing their area descriptions, modalities, spectral-spatial dimensions, and class distributions. To underscore the generalizability of the proposed S$^2$Fin, this table also reports the overall accuracy (\%) of the top-performing baseline method for each dataset alongside our results. The column ‘$\Delta$’ represents the absolute accuracy improvement, demonstrating the consistent superiority of S$^2$Fin across diverse sensor combinations. For brevity, detailed data descriptions, pseudo-color visualizations, and extensive qualitative classification maps are provided in the Supplementary Material B.
	
	\subsection{Experimental Setup}
	The experimental framework is established using PyTorch, executed on an NVIDIA GeForce RTX 3090 24 GB graphics card. All multimodal datasets used are established benchmark datasets and have undergone normalized pixel-level pairing and preprocessing, including min-max normalization and edge-based padding. The optimization strategy adopted is the adaptive moment estimation (Adam) algorithm, with a learning rate set to $5 \times 10^{-4}$ and a weight decay of $4 \times 10^{-4}$. The learning rate modulation is governed by “MultiStepLR” with a decay factor 0.5. We select different local window sizes for different datasets to control the spatial size of the multimodal input, while unifying the size of all spectral patches to 3$\times$3. Furthermore, the trade-off factor $\alpha$ is assigned the value of 0.2. For parameter tuning with a small number of samples, we follow a 5-fold cross-validation within the labeled pool, meaning that for every 10 valid samples, 8 are randomly selected for training and 2 for validation. All Mamba blocks are bidirectional with a depth of two. The embedded features have length 64, and training is performed for 320 epochs. HFEST includes two trainable scalar parameters: a frequency cutoff $f_{\text{cutoff}}$=0.5 and a gain coefficient $g_{\text{amp}}$=0.05, respectively. AFCM follows the 0.5 ratio, with the highest 25\% for augmentation and the lowest 25\% for structure sharing. The trade-off $\gamma$ from SSAF is initialized as 0.5. These parameters are optimized automatically during network training. Scaling factor $k$ is 100 and temperature coefficient $T$ is 0.01. All the experiments are performed 10 times with seeds 0-9. In the following comparative experiments, all four datasets use 10 samples per class to represent a condition of few-sample training. For detailed experiments (datasets, preprocessing, patch extraction, training protocol, and hyperparameters), please refer to the Supplementary Material C and project code repository \footnote{https://github.com/HaoLiu-XDU/SSFin}.
	
	It is worth noting that low- and high-frequency components are defined relatively rather than by fixed absolute indices, so the same rule applies across different datasets and feature-map sizes. The precise index sets and coefficient-selection rules are provided in Supplementary Material D.
	
	In our experiment, we employ four metrics to quantitatively evaluate the classification performance: class-specific accuracy, overall accuracy (OA), average accuracy (AA), and kappa coefficient (Kappa). These metrics provide comprehensive measures of the classification accuracy.
	
	\subsection{Parameter Tuning}
	
	The experiments are constructed to analyze the role of main parameters within the S$^2$Fin model. These parameters are local window size and $\alpha$ in Eq. ~\eqref{3-2-12}, both of which reflect the impact of spatial information on the model. The local window size represents the range of spatial information that the network can perceive, while $\alpha$ is a trade-off parameter that determines the spatial amplitude enhancement. To explore the impact of these parameters on the model, we conducted a series of comparative experiments. Specifically, $\alpha$ and local window size are selected from two sets of values \{0.2, 0.4, 0.6, 0.8, 1.0\} and \{7, 9, 11, 13\}, respectively.
	\begin{table}[hptb]
		\centering
	\begin{minipage}{0.48\textwidth}
		\centering
		\caption{OA (\%) with different parameters for $\alpha$ on the four considered datasets}
		\label{pt1}
		\resizebox{\textwidth}{!}{
			\begin{tabular}{c|c|c|c|c|c|c}
				\hline
				\hline
				Dataset & 0.2 & 0.4 & 0.6 & 0.8 & 1.0 &NSI\\
				\hline
				Houston 2013 & \pmb{89.19} & 89.02 & 88.81 & 88.97 & 88.56&0.0049 \\
				Augsburg & 79.76 & \pmb{79.91} & 79.23 & 79.17 & 79.35 &0.0462\\
				Yellow River Estuary & \pmb{67.54} & 67.25 & 67.18 & 66.99 & 66.73&0.0265 \\
				LCZ HK & \pmb{72.26} & 72.03 & 71.80 & 71.98 & 71.76 &0.0180\\
				\hline
				\hline
		\end{tabular}}
	\end{minipage}
	\hfill
	\begin{minipage}{0.5\textwidth}
		\centering
		\caption{OA (\%) with different parameters for local window size on the four considered datasets}
		\label{pt2}
		\resizebox{0.85\textwidth}{!}{
			\begin{tabular}{c|c|c|c|c|c}
				\hline
				\hline
				Dataset & 7 & 9 & 11 & 13&NSI \\
				\hline
				Houston 2013 & 89.10 & 88.75 & \pmb{89.19} & 89.02&0.0071 \\
				Augsburg & \pmb{79.91} & 78.66 & 77.47 & 76.28 &0.0093\\
				Yellow River Estuary & 65.78 & 66.29 & 66.56 & \pmb{67.54}&0.0120 \\
				LCZ HK & 72.09 & 71.54 & \pmb{72.26} & 70.98 &0.0069\\
				\hline
				\hline
		\end{tabular}}
	\end{minipage}
	
	\end{table}

	Tables \ref{pt1}-\ref{pt2} illustrate the impact of parameters on the model's performance. Observations from the table reveal that a relatively small value of $\alpha$=0.2 optimizes performance. On the other hand, different datasets have different sensitivities to the local window size. We also add the Normalized Sensitivity Index (NSI) to evaluate the robustness of these parameters, showing that the model is more sensitive to $\alpha$.

	\subsection{Ablation Study}
	\begin{table}[hptb]\
		\centering
		\caption{\\OA (\%) obtained in the ablation study on the four considered datasets}
		\label{Table_ablation}
		\resizebox{0.5\textwidth}{!}{{
				\begin{tabular}{c|c|c|c|c|c} 
					\hline 
					\hline
					Dataset&AFCM & HFGM &HFEST& SSAF& S$^2$Fin \\
					\hline
					Houston 2013  & 88.56 & 88.41   &   88.85&   89.02   &\pmb{89.19}     \\
					Yellow River Estuary  & 67.02&66.54&66.96&  67.00 &\pmb{67.56}  \\
					Augsburg  & 78.34&77.83&   79.88&  78.46 &\pmb{79.91}  \\
					LCZ HK  & 71.20 &   71.26 &71.60&   72.12 & \pmb{72.26}  \\
					
					\hline 
					\hline

				\end{tabular}
		}}
	\end{table}
	To assess the effectiveness of the S$^2$Fin framework, we conduct ablation experiments by systematically removing key modules, including AFCM, HFGM, HFEST, and SSAF. The AFCM employs cosine transformation to enhance high-frequency signals while preserving low-frequency components. The HFGM enhances high-frequency amplitudes to enrich detailed information while the HFEST integrates spectral information from HSI or MSI with spatial features for classification. Lastly, the SSAF module refines the fusion of spatial and spectral features post-frequency processing. The respective experiments in Table \ref{Table_ablation} are labeled as “AFCM”, “HFGM”, “HFEST” and “SSAF”.
	
	The experimental results are presented in Table \ref{Table_ablation}. In general, removing the spatial-frequency fusion blocks (AFCM and HFGM) leads to lower OA values across all four datasets, indicating their significance to the model. On the other hand, removing the spatial-spectral fusion block (SSAF) has the least impact on classification performance compared to eliminating other frequency domain components.
	\subsection{Quantitative Results}
	
	\begin{table}[hptb] 
		\centering  
		\caption{Classification results (\%) on the Houston2013 dataset with 10 training samples for each class (bold values are the best and underline values are the second)}  
		\label{Table8}
		\resizebox{\columnwidth}{!}{
			\begin{tabular}{lc|ccccccccc}
	
				\toprule
				\toprule 
				Class  &Numbers & FusAtNet &AsyFFNet&Fusion-HCT  & MACN & CALC &UACL& NCGLF  &MSFMamba& S$^2$Fin\\
				\midrule 
				1 Health Grass         & 1251 &80.10$\pm$3.69&	80.10$\pm$4.18&	82.11$\pm$2.61&	96.29$\pm$1.98&\underline{97.99}$\pm$\underline{1.32}&85.17$\pm$2.43&	96.48$\pm$1.08&92.25$\pm$5.58
				&	\pmb{98.20}$\pm$\pmb{1.73}\\
				2 Stressed Grass   & 1254&	85.29$\pm$5.18&	95.82$\pm$2.03&	97.11$\pm$1.66&\underline{97.67}$\pm$0.98&	87.22$\pm$3.46&\pmb{98.07}$\pm$\pmb{1.04}	&82.30$\pm$4.54&94.28$\pm$3.21
				&	90.56$\pm$4.30\\
				3 Synthetic Grass       & 697& 	83.11$\pm$1.21&	93.30$\pm$0.54&	98.84$\pm$0.12&	99.61$\pm$0.05&	99.13$\pm$0.38&99.35$\pm$0.47&	\underline{99.57}$\pm$\underline{0.25}&99.56$\pm$0.14
				&	\pmb{99.65}$\pm$\pmb{0.25}\\
				4 Tress       & 12444 & 	87.12$\pm$3.65&	88.01$\pm$4.72&	93.44$\pm$3.21&	96.90$\pm$1.38&	92.79$\pm$1.02&\underline{98.14}$\pm$\underline{1.53}&	95.34$\pm$2.44&\pmb{98.94}$\pm$\pmb{3.85}
				&	93.95$\pm$2.33\\
				5 Soil     & 1242&	\underline{99.92}$\pm$\underline{0.02}&	\pmb{100.00}$\pm$\pmb{0.00}&	99.35$\pm$0.04&	97.08$\pm$1.47&	\pmb{100.00}$\pm$\pmb{0.00}&99.92$\pm$0.51&	99.84$\pm$0.05&99.22$\pm$1.86
				&	99.92$\pm$0.12\\
				
				6 Water      & 325&	84.13$\pm$1.99&	81.90$\pm$2.53&	\pmb{100.00}$\pm$\pmb{0.00}&	98.10$\pm$1.07&	82.54$\pm$3.61&\underline{99.68}$\pm$\underline{0.44}&	84.62$\pm$1.77&	\pmb{100.00}$\pm$\pmb{0.00}
				&97.71$\pm$1.32\\
				
				7 Residential        & 1268&	83.70$\pm$3.63&	72.97$\pm$4.28&\underline{93.64}$\pm$\underline{3.45}&	85.37$\pm$2.22&	91.02$\pm$3.70&\pmb{98.73}$\pm$\pmb{0.62}&	81.07$\pm$2.33&90.84$\pm$4.54
				&	87.97$\pm$4.02\\
				8 Commercial         &1244&	67.75$\pm$7.21&	62.64$\pm$4.09&	56.16$\pm$5.25&	62.07$\pm$6.09&	67.75$\pm$5.12&58.91$\pm$6.21&	70.74$\pm$6.56&\pmb{83.65}$\pm$\pmb{6.93}
				&\underline{71.73}$\pm$5.36\\
				
				9 Road   & 1252&	\underline{81.48}$\pm$\underline{2.55}	&62.88$\pm$5.66	&66.26$\pm$3.80&72.46$\pm$3.42&	78.58$\pm$2.52&\pmb{88.65}$\pm$\pmb{3.17}&	78.51$\pm$7.51&80.47$\pm$5.26
				&	74.85$\pm$3.09\\
				
				10 Highway        & 1227 &40.02$\pm$8.92&	55.46$\pm$7.37	&77.49$\pm$5.93&	\underline{78.88}$\pm$\underline{5.24}&	75.35$\pm$4.20&75.84$\pm$2.54	&\pmb{86.88}$\pm$\pmb{6.34}&62.93$\pm$5.53
				&	77.11$\pm$5.87\\
				
				11 Railway        & 1235&	87.51$\pm$3.26&	\pmb{94.61}$\pm$\pmb{2.65}&\pmb{94.61}$\pm$\pmb{3.78}&	\underline{94.12}\underline{4.10}&	72.33$\pm$5.74&88.90$\pm$3.62&	94.09$\pm$3.01&88.39$\pm$4.36
				&92.65$\pm$4.47\\
				
				12 Parking Lot 1     & 1233&	31.81$\pm$14.82&	79.31$\pm$5.51&	\underline{87.00}$\pm$\underline{4.21}&	73.02$\pm$3.85&	68.77$\pm$6.08&49.80$\pm$9.21&	75.67$\pm$6.61&48.85$\pm$5.46
				&	\pmb{87.36}$\pm$\pmb{4.49}\\
				
				13 Parking Lot 2      & 469&	89.32$\pm$2.56&	55.34$\pm$10.78&	\pmb{100.00}$\pm$\pmb{0.00}&	\underline{95.64}$\pm$\underline{2.37}&	82.79$\pm$3.56&84.75$\pm$2.86&	93.18$\pm$3.83&95.18$\pm$3.02
				&	89.76$\pm$1.52\\
				
				14 Tennis Court    & 428&	\pmb{100.00}$\pm$\pmb{0.00}&	\pmb{100.00}$\pm$\pmb{0.00}&	\underline{99.76}$\pm$\underline{0.08}&	95.22$\pm$1.87&	95.93$\pm$0.83&\pmb{100.00}$\pm$\pmb{0.00}&97.43$\pm$0.74&	\pmb{100.00}$\pm$\pmb{0.00}
				&	\pmb{100.00}$\pm$\pmb{0.00}\\
				15 Running Track      & 660&	91.69$\pm$4.50&	\pmb{100.00}$\pm$\pmb{0.00}&	\pmb{100.00}$\pm$\pmb{0.00}&	\pmb{100.00}$\pm$\pmb{0.00}&	99.69$\pm$0.07&99.23$\pm$0.44&	\pmb{100.00}$\pm$\pmb{0.00}&	\pmb{100.00}$\pm$\pmb{0.00}
				&\underline{99.88}$\pm$\underline{0.18}	\\
				\midrule
				&OA   &	77.09$\pm$1.45&	80.66$\pm$1.65&	87.26$\pm$1.52&	87.54$\pm$1.02&	85.01$\pm$1.63&86.42$\pm$0.99&	\underline{87.83}$\pm$\underline{0.76}&86.64$\pm$1.21
				&	\pmb{89.19}$\pm$\pmb{1.06}
				\\
				&AA   &	79.53$\pm$1.28&	81.49$\pm$1.30&	\underline{89.72}$\pm$\underline{1.28}&	89.48$\pm$0.81&	86.11$\pm$1.29&88.34$\pm$0.75&	89.05$\pm$0.60&89.02$\pm$1.07
				&	\pmb{90.75}$\pm$\pmb{0.92}
				\\
				&Kappa&75.24$\pm$1.52&	79.08$\pm$1.88&	86.26$\pm$1.89&	86.54$\pm$1.42&	83.79$\pm$1.54&85.33$\pm$1.04&	\underline{86.84}$\pm$\underline{0.82}&85.57$\pm$1.09
				&	\pmb{88.31}$\pm$\pmb{1.15}
				\\
				\bottomrule 
				\bottomrule

			\end{tabular}
		}
	\end{table}

	\begin{table}[hptb]
		\renewcommand\arraystretch{1}
		\centering  
		\caption{Classification results (\%) on the Augsburg dataset with 10 training samples for each class (bold values are the best and underline values are the second)} 
		\label{Table-h}
		\resizebox{\columnwidth}{!}{
			\begin{tabular}{lc|ccccccccc}
				\toprule
				\toprule 
				Class  &Numbers & FusAtNet& AsyFFNet &Fusion-HCT  & MACN &  CALC &UACL& NCGLF  &MSFMamba& S$^2$Fin\\
				\midrule
				1 Forest & 13507  &	97.79$\pm$1.83&	91.72$\pm$2.14&	92.52$\pm$3.00&	\underline{97.83}$\pm$\underline{1.24}&	97.12$\pm$2.51&94.95$\pm$2.72&	95.30$\pm$1.21&96.82$\pm$2.57
				&	\pmb{98.82}$\pm$\pmb{1.59}
				\\
				2 Residential Area    & 30329&	\pmb{80.95}$\pm$\pmb{3.52}&	78.40$\pm$2.99&	79.01$\pm$4.08&	74.19$\pm$5.67&	\underline{79.04}$\pm$\underline{2.54}&73.81$\pm$3.60&	69.89$\pm$4.79&66.22$\pm$3.42
				&	74.61$\pm$3.94
				\\
				3 Industrial Area     & 3851 & 24.99$\pm$10.62&	53.44$\pm$5.25&	40.87$\pm$6.71&	\underline{60.90}$\pm$\underline{5.03}&	12.05$\pm$8.68	&42.65$\pm$7.22&53.91$\pm$5.32&\pmb{65.36}$\pm$\pmb{5.89}
				&60.20$\pm$3.70
				\\
				4 Low Plants          & 26857 &67.12$\pm$5.32&	68.00$\pm$6.10&	70.75$\pm$3.09&	75.99$\pm$3.51&	77.93$\pm$2.14&79.69$\pm$3.99&	82.34$\pm$4.28&\pmb{84.36}$\pm$\pmb{4.05}
				&	\underline{82.43}$\pm$\underline{3.42}
				\\
				5 Allotment           & 575  & 76.46$\pm$3.00&	86.90$\pm$2.13&	90.27$\pm$2.06&	\underline{96.70}$\pm$\underline{1.17}&	18.76$\pm$8.09&93.45$\pm$1.78&	86.61$\pm$2.78&88.81$\pm$3.45
				&	\pmb{97.03}$\pm$\pmb{1.23}
				\\
				6 Commercial Area     & 1645 & \underline{66.79}$\pm$\underline{2.91}&	51.44$\pm$3.37&	\pmb{68.99}$\pm$\pmb{3.08}&	55.66$\pm$4.03&	39.51$\pm$6.12&49.36$\pm$4.40&	42.25$\pm$8.14&32.95$\pm$5.81
				&	36.27$\pm$4.66
				\\
				7 Water               & 1530  & 38.03+12.79	&\pmb{76.78}$\pm$\pmb{3.22}&	62.04$\pm$4.02&	56.38$\pm$5.31&	50.92$\pm$4.27&\underline{72.89}$\pm$\underline{3.01}&	63.20$\pm$4.24&61.53$\pm$4.57
				&	63.67$\pm$3.85
				\\
				
				\midrule
				&OA  &75.20$\pm$4.76&	75.37$\pm$3.28&	76.18$\pm$5.23&	\underline{77.67}$\pm$\underline{5.58}&	76.68$\pm$6.11&77.56$\pm$3.54&	77.34$\pm$3.93&77.06$\pm$2.36
				&	\pmb{79.91}$\pm$\pmb{1.59}
				\\
				&AA  & 	64.59$\pm$3.28&	72.39$\pm$3.04&	72.06$\pm$4.12	&\pmb{73.95}$\pm$\pmb{3.78}	&53.62$\pm$5.23&72.40$\pm$1.14&	70.50$\pm$3.30&70.86$\pm$1.25
				&	\underline{73.29}$\pm$\underline{0.64}
				\\
				&Kappa& 67.03$\pm$4.90&	67.53$\pm$3.61&	67.91$\pm$5.01	&\underline{70.45}$\pm$\underline{4.99}	&66.87$\pm$6.20&69.96$\pm$3.25&	70.09$\pm$3.80&69.67$\pm$2.33
				&	\pmb{72.96}$\pm$\pmb{1.94}
				\\
				\bottomrule 
				\bottomrule 
			\end{tabular}
		}
	\end{table}
	
	\begin{table}[hptb]
		\centering  
		\caption{Classification results (\%) on the Yellow River Estuary dataset with 10 training samples for each class (bold values are the best and underline values are the second)}  
		\label{Table9}
		\resizebox{\columnwidth}{!}{
			\begin{tabular}{lc|ccccccccc} 
				\toprule
				\toprule 
				Class  &Numbers&  FusAtNet& AsyFFNet&Fusion-HCT  & MACN &  CALC &UACL& NCGLF  &MSFMamba& S$^2$Fin\\
				\midrule
				1 Spartina Alterniflora           & 39784&	63.22$\pm$3.20&	\underline{87.85}$\pm$\underline{1.76}&	81.07$\pm$2.56&	75.35$\pm$2.63&	68.91$\pm$1.79&84.78$\pm$2.03&	87.53$\pm$3.43&\pmb{90.58}$\pm$\pmb{1.23}
				&	75.71$\pm$2.43
				\\
				2  Suaeda Salsa    & 118213& 	49.94$\pm$5.07&	53.16$\pm$3.25&	56.53$\pm$2.16&	59.59$\pm$4.60&	\pmb{85.84}$\pm$\pmb{2.37}&62.74$\pm$3.11&	56.78$\pm$3.48&\underline{65.90}$\pm$\underline{4.26}
				&	63.45$\pm$4.67
				\\
				3 Tamarix Forest     & 35216 & 	\underline{76.63}$\pm$\underline{4.04}&	59.18$\pm$3.81&	65.15$\pm$4.97&	54.24$\pm$3.73&	25.13$\pm$10.45&53.37$\pm$4.30&	\pmb{77.02}$\pm$\pmb{3.84}&46.38$\pm$7.61
				&	72.64$\pm$5.21
				\\
				4  Tidal Creek          & 15673& 59.00$\pm$4.56&	54.40$\pm$3.97&	\underline{74.85}$\pm$\underline{2.75}&	52.22$\pm$4.31&	53.41$\pm$4.08&48.08$\pm$5.15&	\pmb{77.60}$\pm$\pmb{3.30}&66.67$\pm$3.45
				&	73.52$\pm$2.54
				\\
				5  Mudflat                & 24592& 	57.00$\pm$6.49&	48.38$\pm$4.87&	45.72$\pm$5.19&	\pmb{75.49}$\pm$\pmb{3.28}&	19.12$\pm$11.73&\underline{67.40}$\pm$\underline{5.82}&	41.66$\pm$3.71&48.10$\pm$7.21
				&	62.89$\pm$6.37
				\\			
				\midrule
				 &OA& 	57.53$\pm$2.56&	59.56$\pm$2.03&	62.10$\pm$3.16&	62.65$\pm$2.23&	64.60$\pm$5.01&64.59$\pm$1.80&	64.88$\pm$1.55&\underline{65.34}$\pm$\underline{1.96}
				&	\pmb{67.54}$\pm$\pmb{2.21}
				\\
				& AA& 61.09$\pm$2.18&	60.59$\pm$1.53&	64.66$\pm$1.72&	63.38$\pm$1.44&	50.48$\pm$3.80&63.28$\pm$1.93&	\underline{68.12}$\pm$\underline{1.12}&63.52$\pm$2.10
				&	\pmb{69.64}$\pm$\pmb{1.97}
				\\
				&Kappa&	44.26$\pm$2.71&	45.87$\pm$2.51&	49.20$\pm$3.28&	49.37$\pm$3.09&	43.72$\pm$4.65&51.24$\pm$2.37&	\underline{53.02}$\pm$\underline{1.65}&51.76$\pm$2.48
				&	\pmb{55.86}$\pm$\pmb{2.52}
				\\
				\bottomrule 
				\bottomrule 
				
			\end{tabular}
		}
	\end{table}
	
	\begin{table}[hptb]
		\centering  
		\caption{Classification results (\%) on the LCZ HK  dataset with 10 training samples for each class (bold values are the best and underline values are the second)}
		\label{Table10}
		\resizebox{\columnwidth}{!}{
			\begin{tabular}{lc|ccccccccc} 
				\toprule 
				\toprule 
				Class  &Numbers&  FusAtNet& AsyFFNet &Fusion-HCT  & MACN &  CALC&UACL & NCGLF  &MSFMamba& S$^2$Fin\\
				\midrule
				1 Compact High-rise & 631 & 	\pmb{56.52}$\pm$3.58&	40.42$\pm$4.85&	41.71$\pm$5.26&	12.72$\pm$13.69	&50.08$\pm$4.38&45.73$\pm$6.55&	32.69$\pm$13.69&18.52$\pm$10.76
				&	\underline{52.82}$\pm$\underline{3.99}
				\\
				2 Compact Mid-rise              & 179  &72.78$\pm$3.10&	63.31$\pm$4.82&	\underline{74.26}$\pm$\underline{2.17}&	57.40$\pm$3.64&	73.37$\pm$2.10&68.05$\pm$3.88&	61.54$\pm$14.28&72.19$\pm$5.73
				&	\pmb{76.57}$\pm$\pmb{2.32}
				\\
				3 Compact Low-rise & 326& 85.44$\pm$4.57	&\pmb{93.67}$\pm$\pmb{1.36}&	74.05$\pm$3.98&	75.63$\pm$4.22&	\underline{92.41}$\pm$\underline{2.50}&67.41$\pm$5.81&	79.43$\pm$6.23&87.97$\pm$2.26
				&80.76$\pm$7.03
				\\
				4 Open High-rise           & 673 &	35.75$\pm$11.52	&54.90$\pm$6.45	&51.89$\pm$7.33	&\pmb{56.86}$\pm$\pmb{4.92}&	12.07$\pm$15.76&52.19$\pm$7.50&	\underline{55.35}$\pm$\underline{8.51}&41.48$\pm$5.36
				&	38.58$\pm$8.34
				\\
				5  Open Mid-rise & 126& 	50.00$\pm$14.21&	58.62$\pm$10.95&	\pmb{73.28}$\pm$\pmb{6.30}	&\underline{62.93}$\pm$\underline{5.84}&	18.10$\pm$22.86&44.83$\pm$6.73&	34.48$\pm$20.04&43.10$\pm$9.53
				&	52.93$\pm$10.24
				\\
				6 Open Low-rise & 120  & 	56.36$\pm$5.12&	48.18$\pm$6.37&	60.00$\pm$5.45&	49.09$\pm$8.97&	45.45$\pm$6.74&\underline{63.09}$\pm$4.80&	38.18$\pm$10.32&59.09$\pm$9.41
				&	\pmb{66.36}$\pm$\pmb{5.92}
				\\
				7 Large Low-rise  & 137 & 	63.78$\pm$8.33&	40.94$\pm$13.58&	69.29$\pm$6.47&	\underline{72.44}$\pm$\underline{6.93}	&62.99$\pm$4.55	&65.35$\pm$6.82&\pmb{77.17}$\pm$\pmb{7.49}&25.98$\pm$5.29
				&	32.44$\pm$8.05
				\\
				8 Heavy Industry       & 219 & 	\underline{71.77}$\pm$\underline{8.93}&	28.71$\pm$18.52&	46.89$\pm$6.80&	45.93$\pm$10.37&	\pmb{100.00}$\pm$\pmb{0.00}&69.86$\pm$5.93&	64.11$\pm$15.72&55.50$\pm$5.84
				&66.70$\pm$7.71
				\\
				9 Dense Trees & 1616 & 91.34$\pm${2.15}&	87.80$\pm$2.93&	83.50$\pm$4.67&	\pmb{95.39}$\pm$3.52	&\underline{94.71}$\pm$\underline{2.31}	&69.42$\pm$6.57&88.79$\pm$6.43&90.54$\pm$2.30
				&	86.66$\pm$3.22
				\\
				10 Scattered Trees          & 540 &54.72$\pm$8.16	&24.72$\pm$20.30	&65.28$\pm$9.65&	32.08$\pm$16.46&	\underline{72.26}$\pm$\underline{5.36}&\pmb{77.55}$\pm$\pmb{6.02}&	55.28$\pm$12.30&55.28$\pm$6.31
				&	66.64$\pm$6.98
				\\
				11 Bush and Scrub         & 691  & 	53.30$\pm$7.49&	64.17$\pm$6.38	&\pmb{94.27}$\pm$\pmb{2.50}&	62.85$\pm$7.44&	54.04$\pm$9.01&54.63$\pm$8.16&	\underline{79.30}$\pm$\underline{7.65}&58.52$\pm$7.28
				&69.54$\pm$6.86
				\\
				12 Low Plants          & 985 & 	36.36$\pm$12.78&	37.03$\pm$14.60&	17.85$\pm$21.68&	40.16$\pm$8.07&	20.72$\pm$18.52&40.23$\pm$8.55&	35.08$\pm$15.73&\pmb{46.56}$\pm$\pmb{5.72}
				&	\underline{40.82}$\pm$\underline{3.52}
				\\
				13 Water        & 2603 & 	68.11$\pm$4.51&	90.78$\pm$2.35&	\underline{94.91}$\pm$\underline{1.87}&	94.99$\pm$2.03&	96.14$\pm$3.17&\pmb{97.69}$\pm$\pmb{0.93}	&91.86$\pm$1.64&89.86$\pm$2.73
				&	92.48$\pm$10.58
				\\
				\midrule
				&OA & 	63.94$\pm$4.37&	68.20$\pm$3.65&	\underline{71.87}$\pm$\underline{3.42}&	70.11$\pm$3.91&	69.24$\pm$2.13&70.34$\pm$2.50&	71.39$\pm$2.63&68.66$\pm$2.24
				&	\pmb{72.26}$\pm$\pmb{2.75}
				\\
				&AA &61.43$\pm$2.08&	56.40$\pm$3.17&	\pmb{65.02}$\pm$\pmb{2.97}&	58.50$\pm$3.20&	59.75$\pm$1.85&62.77$\pm$2.19&	61.02$\pm$3.63&57.28$\pm$2.03
				&	\underline{63.33}$\pm$\underline{1.37}
				\\
				&Kappa&	59.05$\pm$4.80&	62.15$\pm$3.84&	\underline{67.06}$\pm$\underline{3.59}&	64.73$\pm$4.26&	63.80$\pm$2.24&65.33$\pm$2.67&	66.45$\pm$2.96&63.27$\pm$2.26
				&\pmb{67.42}$\pm$\pmb{2.83}
				
				\\
				\bottomrule 
				\bottomrule
				
			\end{tabular}
		}
	\end{table}
	
	To illustrate the effectiveness of the proposed S$^2$Fin, we have conducted a comparative analysis with seven state-of-the-art multimodal classification models. FusAtNet utilizes a self-attention mechanism to extract spectral features and employs a cross-modality attention mechanism to extract spatial features from multimodal data for land-cover classification. AsyFFNet has crafted an asymmetric neural network with weight-sharing residual blocks for multimodal feature extraction and introduced a channel exchange mechanism and sparse constraints for feature fusion. Furthermore, we have selected five methods that concentrate on global and local multimodal features. Fusion-HCT and MACN integrate CNNs and transformers to capture both local and global features, introducing innovative attention mechanisms for multimodal feature fusion.  CALC fuses high-order semantic and complementary information for accurate classification. UACL is based on a contrastive learning strategy to access reliable multimodal samples. NCGLF enhances CNN and transformer structures with structural information learning and invertible neural networks. MSFMamba utilizes multiscale feature fusion state space model to extract multisource information. The hyperparameters of the comparative experiments followed those in the original paper, and the same random seeds are used. The performance of these methods is summarized in Tables \ref{Table8}-\ref{Table10}. The following conclusions can be inferred.
	
	\begin{enumerate}
		
		\item Overall, advanced approaches which prioritize the integration of global and local features for multimodal data fusion demonstrate excellent classification performance. These approaches tend to outperform those that focus solely on attention mechanisms and network architectures. Meanwhile, these methods exhibit consistent performance across various datasets, attributed to their diverse strategies for fusing global and local information.
		
		\item Leveraging guidance from frequency domain learning, S$^2$Fin has achieved enhanced multimodal feature fusion, reflected in its higher OA, AA, and Kappa scores. Across the four datasets, S$^2$Fin has consistent improvements upon the previous state-of-the-art model by 1.36\%, 2.66\%, 2.24\% and 0.39\% on the OA metric. 
		
		\item S$^2$Fin emphasis on the high-frequency components of multimodal data enables its effective extraction of details information and classification of complex scenes. For example, from the figures and tables, one can see that on the Augsburg dataset, S$^2$Fin has achieved good classification for 3 out of 7 categories. Notably, the “Forest”, "Low Plants" and “Allotment” classes, which are challenging to distinguish due to their similarities, all achieved commendable classification results. Similarly, on the Houston 2013 dataset, S$^2$Fin has the high classification accuracy in 6 out of the 15 categories, with a good performance in similar “Commercial” and "Residential" class over comparison methods.

	\end{enumerate}
	\subsection{Uncertainty and Robustness Analysis}
	To assess the statistical reliability and generalization capability of the proposed S$^2$Fin, we conduct an extensive uncertainty analysis spanning 10 independent experimental runs for each dataset, maintaining random seeds 0-9. As summarized in Table \ref{tab:uncertainty}, we evaluate model uncertainty using standard deviation, Coefficient of Variation (CV), NSI, and the 95\% Confidence Interval (CI) calculated via the $t$-distribution. S$^2$Fin consistently demonstrated very low variance, with the CV remaining below 5\% across all multimodal datasets. Furthermore, to validate the performance advantages over the latest baselines (NCGLF and MSFMamba), we perform paired $t$-tests and computed Cohen’s $d$ effect sizes. The results conclusively demonstrate that S$^2$Fin achieves statistically significant improvements ($p < 0.05$ and $d > 0.8$) in the vast majority of comparisons.
	
\begin{figure}[htbp]
	\centering
	\includegraphics[width=0.9\linewidth]{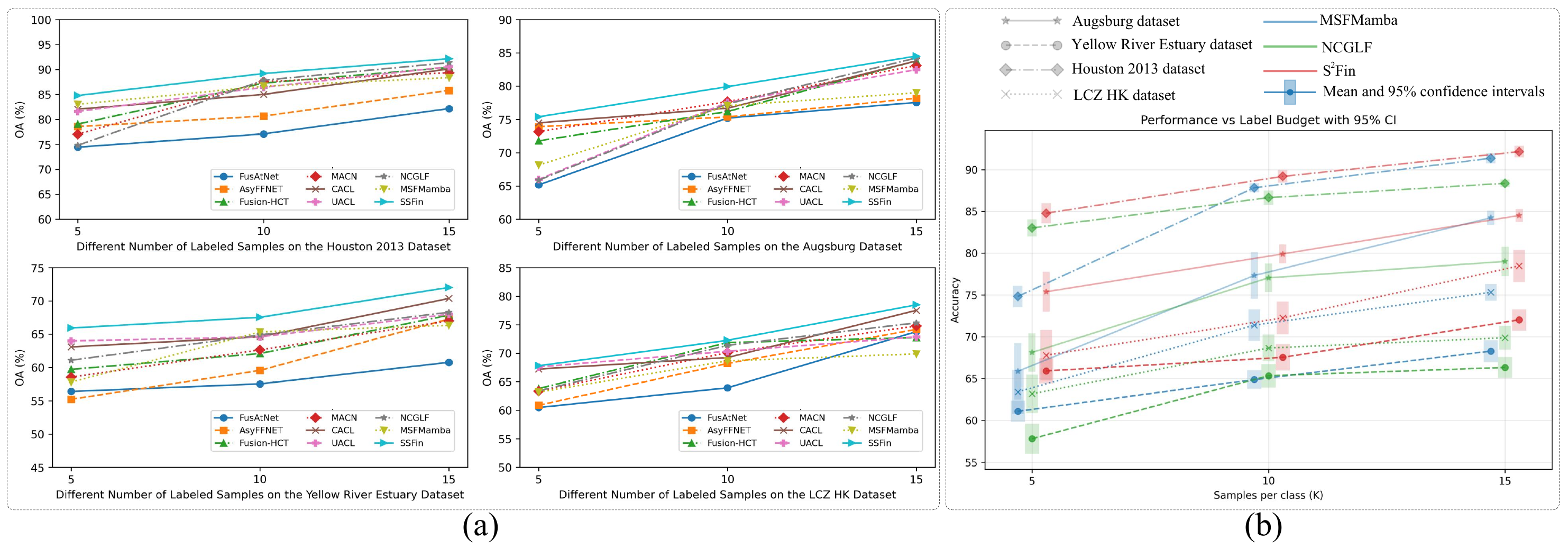}
	\caption{Behaviour of the OA\% versus different number of labeled samples on the four considered datasets. (a) Mean OA performance across all methods. (b) Uncertainty bands (95\% confidence intervals) of S$^2$Fin, NCGLF, and MSFMamba. } 
	\label{fig13}
\end{figure}

	To verify the effect of the model under different numbers of training samples, we conducted experiments with 5, 10 and 15 labeled samples for each class. Fig. \ref{fig13} shows that S$^2$Fin achieves the best OA and exhibits good robustness.
	
	\begin{table*}[t]
		\centering
		\caption{Uncertainty and robustness analysis of the proposed S$^2$Fin across 10 runs. Paired t-tests and Cohen's $d$ are computed against NCGLF ($p$-val1 and Cohen's $d$1) and MSFMamba ($p$-val2 and Cohen's $d$2).}
		\resizebox{0.9\textwidth}{!}{
			\begin{tabular}{c|ccc|ccc|ccc|ccc}
				\toprule
				\toprule
				Metric &
				\multicolumn{3}{c}{Augsburg} &
				\multicolumn{3}{c}{Yellow River Estuary} &
				\multicolumn{3}{c}{Houston 2013} &
				\multicolumn{3}{c}{LCZ HK} \\
				\cline{2-13}
				& OA & AA & Kappa & OA & AA & Kappa & OA & AA & Kappa & OA & AA & Kappa \\
				\hline
				Mean (\%) &79.91&73.29&72.96&67.54&69.64&55.86&89.19&90.75&88.31&72.26&63.33&67.42\\
				Std (\%) &1.59&0.64&1.94&2.21&1.97&2.52&1.06&0.92&1.15&2.75&1.37&2.83\\
				CV (\%) &1.99&0.87&2.66&3.27&2.83&4.51&1.19&1.01&1.30&3.80&2.16&4.20\\
				95\%CI (\%) &$\pm$1.2&$\pm$0.48&$\pm$1.46&$\pm$1.67&$\pm$1.48&$\pm$1.90&$\pm$0.80&$\pm$0.69&$\pm$0.87&$\pm$2.07&$\pm$1.03&$\pm$2.13\\
				NSI  &0.0623&0.0273&0.0843&0.1220&0.0864&0.1466&0.0330&0.0280&0.0371&0.1489&0.0575&0.1646\\
				$p$-val1 & 0.039 & 0.018 & 0.047 & $<$0.001 & 0.008 & 0.005 & $<$0.001 & $<$0.001 & 0.003 & 0.385 & 0.036 & 0.362 \\
				$p$-val2 & 0.018 & $<$0.001 & 0.026 & 0.010& $<$0.001 & 0.024 & $<$0.001 & $<$0.001 & $<$0.001 & 0.002 & $<$0.001 & 0.007 \\
				Cohen's $d$1 & 0.76 & 0.91 & 0.73 & 2.17 & 1.07 & 1.15 & 2.32 & 1.76 & 1.25 & 0.29 & 0.78 & 0.30 \\
				Cohen's $d$2& 0.92&1.51&0.84&1.03&1.86&0.86&1.73&1.96&2.02&1.35&3.34&1.09 \\
				\bottomrule
				\bottomrule
		\end{tabular}}
	\label{tab:uncertainty}
	\end{table*}
	
	\subsection{Qualitative Results}
	\begin{figure}[htbp]
		\centering
		\includegraphics[width=0.9\textwidth]{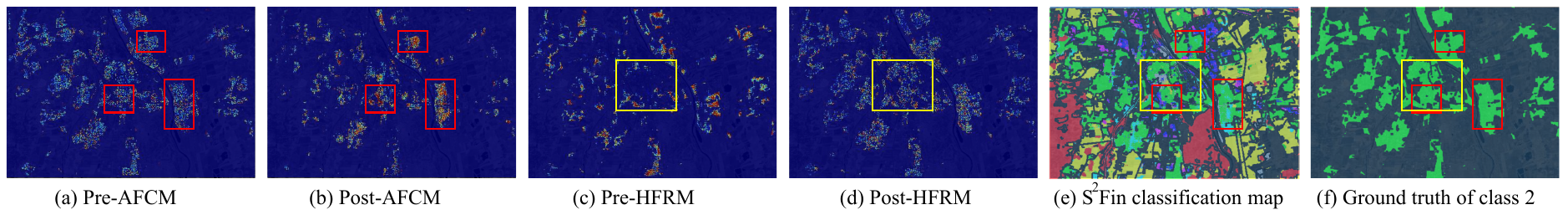}
		
		\caption{Grad-CAM visualization of gradient activation for class 2 residential area of the Augsburg dataset. (a) HSI features before AFCM. (b) HSI features after AFCM. (c) HSI features before HFRM. (d) HSI features after HFRM. (e) S$^2$Fin classification mapping of all categories. (f) Ground-truth map. Colors range from blue (low activation) to red (high activation).} 
		\label{vis}
	\end{figure}
	We apply Grad-CAM to produce class-specific activation maps and visualize how the proposed frequency modules affect attention. Taking Class 2 residential area in the Augsburg dataset as an example, Fig. \ref{vis}(a–d) reports gradient-activation maps before/after the shallow AFCM and deep HFRM, while Fig. \ref{vis}(e)(f) are the corresponding all-class classification maps and ground truth maps for that class. For each patch we compute a Grad-CAM map, concatenate the patch maps into a full-image heatmap and average overlapping locations. The resulting map is normalized to [0,1] after clipping the top 1\% of extreme values. The red boxes in the figure indicate that AFCM enhances gradient attention to previously neglected fine local details. The yellow boxes show that HFRM amplifies previously weaker high-frequency regions of phase coherence, thus restoring regions consistent with the true values. These observations confirm the qualitative improvements in classification plot (e), demonstrating that enhanced frequency perception pays attention to discriminative details.
	
	Note that the classification maps of some baseline methods are depicted in Supplementary Material E for a qualitative comparison. 
	
	\subsection{Cross-Region Generalization Analysis}

	\begin{table*}[htbp]
		\centering
		\caption{Cross-region generalization and transfer learning analysis. Results are shown as "Direct Training" $\rightarrow$ "Transfer Learning" (Pre-trained on source with 10 samples/class).}
		\label{tab:cross_region}
		\resizebox{\textwidth}{!}{
			\begin{tabular}{lc|c|c|c|c|c|c}
				\toprule
				\toprule
				Transfer Case & Metric & 1 Sample & 2 Samples & 3 Samples & 4 Samples & 5 Samples & 10 Samples \\
				\midrule
				\multirow{3}{*}{HK $\rightarrow$ Berlin} & OA (\%) & $47.74 \pm 4.00 \rightarrow 56.71 \pm 5.62$ & $61.57 \pm 2.95 \rightarrow 64.96 \pm 2.60$ & $71.57 \pm 3.26 \rightarrow 72.07 \pm 2.13$ & $74.62 \pm 2.46 \rightarrow 75.80\pm 1.50$ & $75.95 \pm 2.74 \rightarrow 76.24\pm 2.45$ & $79.07 \pm 2.17 \rightarrow 81.55 \pm 1.44$ \\
				& AA (\%) & $48.24 \pm 2.48 \rightarrow 52.61 \pm 2.29$ & $61.09 \pm 1.78 \rightarrow 63.48 \pm 1.86$ & $68.03 \pm 2.51 \rightarrow 67.90 \pm 1.54$ & $71.34 \pm 1.85 \rightarrow 71.80 \pm 1.64$ & $75.64 \pm 2.00 \rightarrow 75.18 \pm 1.35$ & $73.93 \pm 1.46 \rightarrow  81.03 \pm 0.93$ \\
				& Kappa (\%) & $40.76 \pm 4.27 \rightarrow 50.73 \pm 5.86$ & $56.32 \pm 3.12 \rightarrow 59.91 \pm 2.82$ & $67.28 \pm 3.64 \rightarrow 67.87 \pm 2.39$ & $70.95 \pm 2.74 \rightarrow 72.14 \pm 1.68$ & $72.47 \pm 3.02 \rightarrow 72.75 \pm 2.67$ & $74.18 \pm 2.46 \rightarrow 78.79 \pm 1.60$ \\
				\midrule
				\multirow{3}{*}{Berlin $\rightarrow$ HK} & OA (\%) & $58.30 \pm 2.87 \rightarrow 62.45 \pm 2.69$ & $68.37 \pm 2.48 \rightarrow 70.09 \pm 4.74$ & $71.79 \pm 3.68 \rightarrow 72.43 \pm 4.52$ & $72.76 \pm 4.18 \rightarrow 74.17 \pm 4.51$ & $74.27 \pm 3.52 \rightarrow 74.87 \pm 5.22$ & $80.10 \pm 3.33 \rightarrow 81.91 \pm 2.04$ \\
				& AA (\%) & $42.94 \pm 2.10 \rightarrow 43.84 \pm 1.77$ & $56.52 \pm 1.33 \rightarrow 58.96 \pm 2.13$ & $58.96 \pm 2.13 \rightarrow 62.20 \pm 1.78$ & $62.41 \pm 1.51 \rightarrow 67.27 \pm 2.06$ & $67.78 \pm 1.91 \rightarrow 70.47 \pm 2.23$ & $75.58 \pm 1.43 \rightarrow 82.25 \pm 1.44$ \\
				& Kappa (\%)& $49.71 \pm 3.35 \rightarrow 53.86 \pm 3.09$ & $61.47 \pm 2.85 \rightarrow 63.59 \pm 5.17$ & $65.47 \pm 4.10 \rightarrow 66.37 \pm 5.05$ & $66.62 \pm 4.53 \rightarrow 68.41 \pm 5.07$ & $68.53 \pm 3.88 \rightarrow 69.34 \pm 5.88$ & $75.48 \pm 3.76 \rightarrow 79.22 \pm 2.26$ \\
				\bottomrule
				\bottomrule
			\end{tabular}
		}
	\end{table*}
	To evaluate the cross-regional generalization of S$^2$Fin, we employ a transfer learning across different cities, i.e., Berlin and Hong Kong. The former uses the same data source as the LCZ HK dataset, Sentinel 1 and 2 satellites, while they share ten common categories (see Supplementary Material F). Specifically, the model is pre-trained in the source region using 10 labeled samples per category. Then, it is fine-tuned and tested in the target region using different sample sizes ($n \in \{1, 2, 3, 4, 5, 10\}$). As shown in Table \ref{tab:cross_region}, the S$^2$Fin framework exhibits good cross-regional robustness. For example, when the number of labeled samples in the target region is very small (1 or 2 samples per class), transfer learning can significantly improve model performance. The results demonstrate that S$^2$Fin can extract frequency-domain invariant features, enabling the model to adapt to new regions with minimal supervision.
	
	\subsection{Analysis of the Computational Complexity}

	\begin{figure}[htbp]
		\centering
		\includegraphics[width=0.3\linewidth]{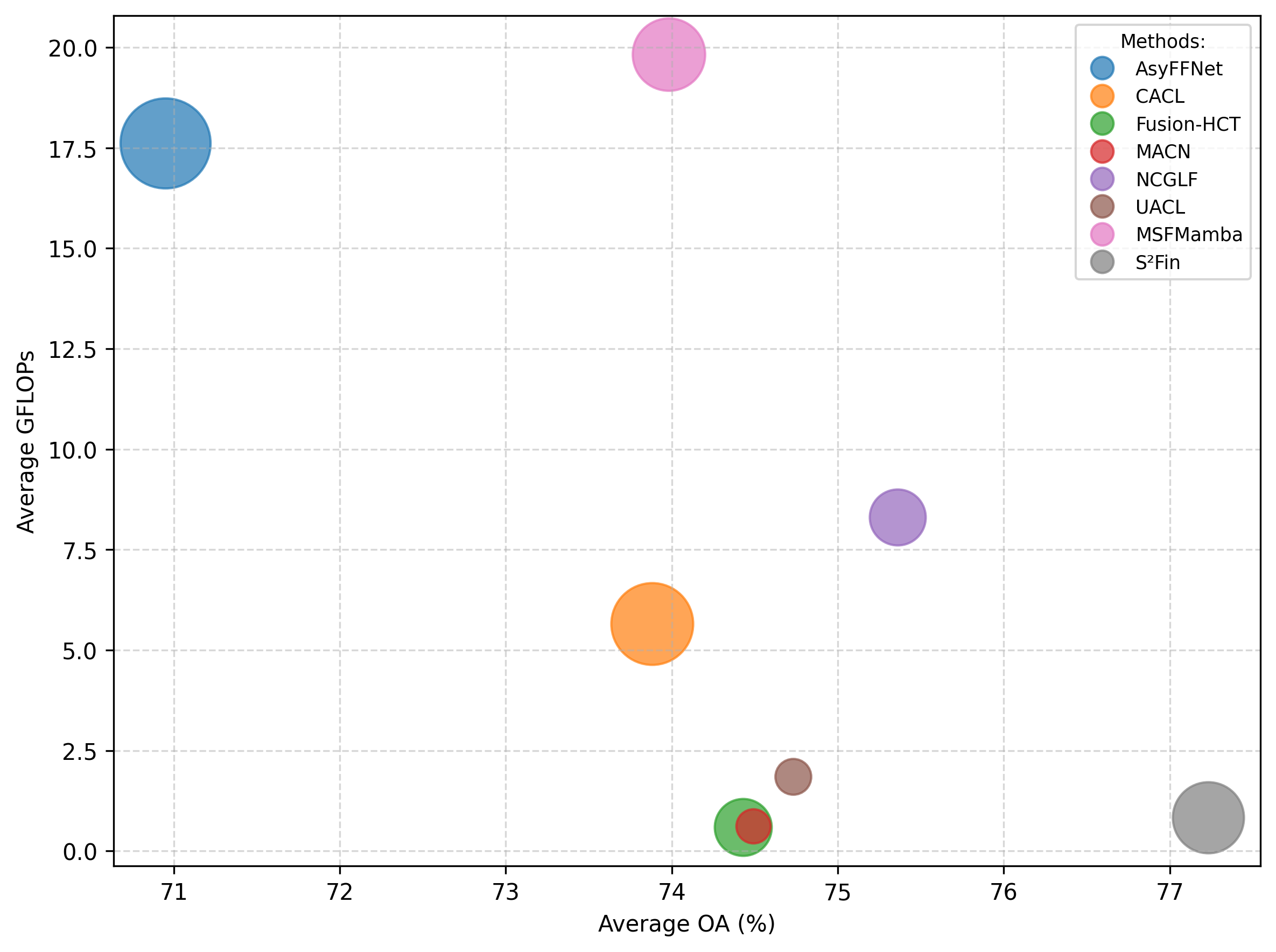}
		
		\caption{The relationship between the average OA and the average computational complexity (GFLOPs) of different methods.} 
		\label{computational_oa}
	\end{figure}
	\begin{table*}[t]
		\centering
		\caption{Number of parameters (M, million) and GFLOPs of considered methods}
		\label{TIME}
		\resizebox{0.75\textwidth}{!}{
			\begin{tabular}{cccccccccc}
				\toprule
				\toprule
				& &  AsyFFNET& CALC&Fusion-HCT&MACN&NCGLF&UACL&	MSFMamba&S$^2$Fin\\
				\hline
				\multirow{2}*{Augsburg}
				& Params. (M) &1.08	&0.94&	0.43&	0.17&	0.44&	0.19&		0.82	&0.63
				\\		
				& GFLOPs  & 17.76&	7.23&	0.59&	0.70&	8.72&	2.38&		25.17	&	0.68
				\\	
				\hline
				\multirow{2}*{Yellow River Estuary}
				& Params. (M) &1.08	&0.92&	0.43&	0.17&	0.44&	0.18&		0.78&		0.70\\
				& GFLOPs  &17.72&	6.80&	0.59&	0.70&	8.72&	2.24&	25.15&		0.99\\
				\hline
				\multirow{2}*{Houston 2013}
				& Params. (M) &1.08&	0.90&	0.43&	0.17&	0.44&	0.18&		0.97	&	0.70
				\\
				& GFLOPs  &17.65&	6.12&	0.59&	0.70&	8.72&	2.01&	25.17	&0.95	
				\\
				\hline
				\multirow{2}*{LCZ HK}
				& Params. (M) &1.06	&0.79&	0.43&	0.07&	0.34&	0.13&	0.21	&	0.65
				\\
				& GFLOPs  &17.32&	2.47&	0.59&	0.37&	7.07&	0.80&		3.83&		0.70
				\\		
				\bottomrule
				\bottomrule
		\end{tabular}}
	\end{table*}
	We evaluate each model’s computational complexity in terms of GFLOPs and parameter count (in millions) in the Table \ref{TIME}. Fig. \ref{computational_oa} shows the relationships between the average (computed on the four datasets) OA and computational complexity (GFLOPs) for the different considered methods. 

	Although the proposed model contains multiple frequency interaction modules, these methods are simple and do not require complex training when embedded in the network, so that the computational cost remains moderate. This is mainly due to the lightweight design of the frequency modules and the compact Mamba backbone. Compared with Mamba-based architectures \cite{10856240}, the proposed method achieves improved classification accuracy while maintaining competitive computational efficiency.
	Furthermore, its number of parameters does not increase significantly with respect to those of other methods and remains lower than those of the AsyFFNET and the CALC. Overall, S$^2$Fin combines low computational complexity with superior performance.

	\section{Conclusion}
	\label{conclusion}
	In this study, we have introduced S$^2$Fin to improve pixel-level, few-sample multimodal remote sensing classification. By using the use of the frequency domain via the HFEST module, the model successfully captures sparse but critical high-frequency details. Our depth-wise spatial frequency fusion strategy (AFCM and HFRM) combines low-frequency structural features with fine high-frequency details. Experimental results across four benchmark multimodal datasets demonstrate that S$^2$Fin consistently achieves superior OA in few-sample scenarios.
	
	The implications of this study lie in the ability of S$^2$Fin to extract high-fidelity features from redundant multimodal signals. From a practical standpoint, the S$^2$Fin architecture is promising for real-time and label-scarce Earth observation tasks, such as rapid disaster response and precise land-cover mapping. Besides, feature alignment and enhancement in the frequency domain provide a new perspective for joint interpretation of multimodal signals and frequency-aware deep learning.

	This study still has some limitations. First, the design of S$^2$Fin relies on attention mechanisms and Mamba modules, but its exploration of classic network architectures and fusion strategies, such as residual networks and UNet architectures, is limited. Second, the experimental analysis was done on four different datasets covering urban, rural, and wetland regions, but it has not been tested in specific or large-scale regions. Finally, although the frequency domain transformation is efficient, the computational overhead in ultra-large-scale deployments may pose scalability challenges.
	
	Future research will focus on the following key areas: 1) Exploring the integration of frequency domain learning paradigms with classic deep learning architectures and large-scale deployment of foundation models. 2) Extending the S$^2$Fin framework to other multimodal tasks and practical applications, such as rapid disaster change detection. 3) Developing reliable, interpretable, and scalable frequency domain learning strategies and combining them with other few-shot learning paradigms to address potential overfitting risks and enhance robustness. Regarding ethical and social implications, the high-precision models must be with a responsible AI framework.

	\biboptions{numbers,sort&compress}
	\bibliographystyle{elsarticle-num}
	\bibliography{bibfile}

@article{qingyun2022cross,
	title={Cross-modality attentive feature fusion for object detection in multispectral remote sensing imagery},
	author={Qingyun, Fang and Zhaokui, Wang},
	journal={Pattern Recognit.},
	volume={130},
	pages={108786},
	year={2022},
	publisher={Elsevier}
}

@article{singh2020new,
	title={A new homomorphic and method noise thresholding based despeckling of SAR image using anisotropic diffusion},
	author={Singh, Prabhishek and Shree, Raj},
	journal={J. King Saud Univ.-Comput. Inf. Sci.},
	volume={32},
	number={1},
	pages={137--148},
	year={2020},
	publisher={Elsevier}
}

@article{singh2021new,
	title={A new SAR image despeckling using correlation based fusion and method noise thresholding},
	author={Singh, Prabhishek and Shree, Raj and Diwakar, Manoj},
	journal={J. King Saud Univ.-Comput. Inf. Sci.},
	volume={33},
	number={3},
	pages={313--328},
	year={2021},
	publisher={Elsevier}
}

@article{hong2021multimodal,
	title={Multimodal remote sensing benchmark datasets for land cover classification with a shared and specific feature learning model},
	author={Hong, Danfeng and Hu, Jingliang and Yao, Jing and Chanussot, Jocelyn and Zhu, Xiao Xiang},
	journal={ISPRS J. Photogramm. Remote Sens.},
	volume={178},
	pages={68--80},
	year={2021},
	publisher={Elsevier}
}

@article{singh2021review,
	title={A Review on SAR Image and its Despeckling},
	author={Singh, Prabhishek and Diwakar, Manoj and Shankar, Achyut and Shree, Raj and Kumar, Manoj},
	journal={Arch. Comput. Methods Eng.},
	volume={28},
	number={7},
	pages={4633--4653},
	year={2021},
	publisher={Springer}
}

@inproceedings{singh2016analysis,
	title={Analysis and effects of speckle noise in SAR images},
	author={Singh, Prabhishek and Shree, Raj},
	booktitle={Proc. 2nd Int. Conf. Adv. Comput., Commun. Autom. (ICACCA-Fall), 2016},
	pages={1--5},
	year={2016},
	organization={IEEE}
}

@article{singh2018new,
	title={A new SAR image despeckling using directional smoothing filter and method noise thresholding},
	author={Singh, Prabhishek and Shree, Raj},
	journal={Eng. Sci. Technol., Int. J.},
	volume={21},
	number={4},
	pages={589--610},
	year={2018},
	publisher={Elsevier}
}

@article{he2017environmental,
	title={Environmental degradation in the urban areas of China: Evidence from multi-source remote sensing data},
	author={He, Chunyang and Gao, Bin and Huang, Qingxu and Ma, Qun and Dou, Yinyin},
	journal={Remote Sens. Environ.},
	volume={193},
	pages={65--75},
	year={2017},
	month={Mar.},
	publisher={Elsevier}
}

@article{liu2025three,
	title={A three-dimensional feature-based fusion strategy for infrared and visible image fusion},
	author={Liu, Xiaowen and Huo, Hongtao and Yang, Xin and Li, Jing},
	journal={Pattern Recognit.},
	volume={157},
	pages={110885},
	year={2025},
	publisher={Elsevier}
}

@article{wang2025lmfnet,
	title={LMFNet: Lightweight Multimodal Fusion Network for high-resolution remote sensing image segmentation},
	author={Wang, Tong and Chen, Guanzhou and Zhang, Xiaodong and Liu, Chenxi and Wang, Jiaqi and Tan, Xiaoliang and Zhou, Wenlin and He, Chanjuan},
	journal={Pattern Recognit.},
	volume={164},
	pages={111579},
	year={2025},
	publisher={Elsevier}
}

@article{ye2025lightweight,
	title={A Lightweight Multilevel Multiscale Dual-Path Fusion Network for Remote Sensing Semantic Segmentation},
	author={Ye, Hong and Chang, Jiaming and Wang, Ke and Jia, Zhaohong and Sun, Wei and Li, Zhiwei},
	journal={Pattern Recognit.},
	pages={112483},
	year={2025},
	publisher={Elsevier}
}

@article{qiao2018joint,
	title={Joint bilateral filtering and spectral similarity-based sparse representation: a generic framework for effective feature extraction and data classification in hyperspectral imaging},
	author={Qiao, Tong and Yang, Zhijing and Ren, Jinchang and Yuen, Peter and Zhao, Huimin and Sun, Genyun and Marshall, Stephen and Benediktsson, Jon Atli},
	journal={Pattern Recognit.},
	volume={77},
	pages={316--328},
	year={2018},
	publisher={Elsevier}
}

@article{xue2025multimodal,
	title={Multimodal self-supervised learning for remote sensing data land cover classification},
	author={Xue, Zhixiang and Yang, Guopeng and Yu, Xuchu and Yu, Anzhu and Guo, Yinggang and Liu, Bing and Zhou, Jianan},
	journal={Pattern Recognit.},
	volume={157},
	pages={110959},
	year={2025},
	publisher={Elsevier}
}

@ARTICLE{10856240,
	author={Gao, Feng and Jin, Xuepeng and Zhou, Xiaowei and Dong, Junyu and Du, Qian},
	journal={IEEE Trans. Geosci. Remote Sens.},  
	title={MSFMamba: Multiscale Feature Fusion State Space Model for Multisource Remote Sensing Image Classification}, 
	year={2025},
	volume={63},
	number={},
	pages={1-16},
	month={Jan.},
}

@article{gao2022fusion,
	title={Fusion classification of HSI and MSI using a spatial-spectral vision transformer for wetland biodiversity estimation},
	author={Gao, Yunhao and Song, Xiukai and Li, Wei and Wang, Jianbu and He, Jianlong and Jiang, Xiangyang and Feng, Yinyin},
	journal={Remote Sens.},
	volume={14},
	number={4},
	pages={850},
	month={Feb.},
	year={2022},
	publisher={MDPI}
}

@ARTICLE{9174822,
	author={Hong, Danfeng and Gao, Lianru and Yokoya, Naoto and Yao, Jing and Chanussot, Jocelyn and Du, Qian and Zhang, Bing},
	journal={IEEE Trans. Geosci. and Remote Sens.}, 
	title={More Diverse Means Better: Multimodal Deep Learning Meets Remote-Sensing Imagery Classification}, 
	year={2021},
	volume={59},
	number={5},
	month={Aug.},
	pages={4340-4354},
}

@article{yu2024mambaout,
	title={Mambaout: Do we really need mamba for vision?},
	author={Yu, Weihao and Wang, Xinchao},
	journal={arXiv preprint arXiv:2405.07992},
	year={2024}
}

@article{liu2024hybrid,
	title={A Hybrid Multi-Task Learning Network for Hyperspectral Image Classification with Few Labels},
	author={Liu, Hao and Zhang, Mingyang and Di, Ziqi and Gong, Maoguo and Gao, Tianqi and Qin, A Kai},
	journal={IEEE Trans. Geosci. Remote Sens.}, 
	year={2024},
	volume={62},
	number={},
	pages={1-16},
	month={Jan.},
}

@article{sun2024unsupervised,
	title={Unsupervised multi-branch network with high-frequency enhancement for image dehazing},
	author={Sun, Hang and Luo, Zhiming and Ren, Dong and Du, Bo and Chang, Laibin and Wan, Jun},
	journal={Pattern Recognit.},
	volume={156},
	pages={110763},
	year={2024},
	month={Dec.},
	publisher={Elsevier}
}

@article{sun2024high,
	title={High-frequency and low-frequency dual-channel graph attention network},
	author={Sun, Yukuan and Duan, Yutai and Ma, Haoran and Li, Yuelong and Wang, Jianming},
	journal={Pattern Recognit.},
	volume={156},
	pages={110795},
	year={2024},
	publisher={Elsevier}
}

@article{song2025efficient,
	title={Efficient frequency feature aggregation transformer for image super-resolution},
	author={Song, Jianwen and Sowmya, Arcot and Sun, Changming},
	journal={Pattern Recognit.},
	pages={111735},
	year={2025},
	publisher={Elsevier}
}

@ARTICLE{9778017,
	author={Behjati, Parichehr and Rodriguez, Pau and Tena, Carles Fernández and Mehri, Armin and Roca, F. Xavier and Ozawa, Seiichi and Gonzàlez, Jordi},
	journal={IEEE Access}, 
	title={Frequency-Based Enhancement Network for Efficient Super-Resolution}, 
	year={2022},
	volume={10},
	number={},
	pages={57383-57397},
	month={May.},
	}

@inproceedings{wang2023learning,
	title={Learning high-frequency feature enhancement and alignment for pan-sharpening},
	author={Wang, Yingying and Lin, Yunlong and Meng, Ge and Fu, Zhenqi and Dong, Yuhang and Fan, Linyu and Yu, Hedeng and Ding, Xinghao and Huang, Yue},
	booktitle={Proc. 31st ACM Int.l Conf. Multimedia},
	pages={358--367},
	year={Oct. 2023}
}

@ARTICLE{10648934,
	author={Chen, Linwei and Fu, Ying and Gu, Lin and Yan, Chenggang and Harada, Tatsuya and Huang, Gao},
	journal={IEEE Trans. Pattern Anal. Mach. Intell.}, 
	title={Frequency-Aware Feature Fusion for Dense Image Prediction}, 
	year={2024},
	volume={46},
	number={12},
	month={Aug.},
	pages={10763-10780},
	}

@article{lu2023coupled,
	title={Coupled adversarial learning for fusion classification of hyperspectral and LiDAR data},
	author={Lu, Ting and Ding, Kexin and Fu, Wei and Li, Shutao and Guo, Anjing},
	journal={Inf. Fusion},
	volume={93},
	month={May},
	pages={118--131},
	year={2023},
	publisher={Elsevier}
}

@ARTICLE{10540387,
	author={Ding, Kexin and Lu, Ting and Li, Shutao},
	journal={IEEE Trans. Geosci. Remote Sens.}, 
	title={Uncertainty-Aware Contrastive Learning for Semi-Supervised Classification of Multimodal Remote Sensing Images}, 
	year={2024},
	volume={62},
	number={},
	month={May},
	pages={1-13},
	}

@ARTICLE{9716784,
	author={Li, Wei and Gao, Yunhao and Zhang, Mengmeng and Tao, Ran and Du, Qian},
	journal={IEEE Trans. Neural Netw. Learn. Syst.}, 
	title={Asymmetric Feature Fusion Network for Hyperspectral and SAR Image Classification}, 
	year={2023},
	volume={34},
	month={Oct.},
	number={10},
	pages={8057-8070},
	}

@ARTICLE{10066307,
	author={Gao, Yunhao and Zhang, Mengmeng and Li, Wei and Song, Xiukai and Jiang, Xiangyang and Ma, Yuanqing},
	journal={IEEE Trans. Geosci. Remote Sens.}, 
	title={Adversarial Complementary Learning for Multisource Remote Sensing Classification}, 
	year={2023},
	volume={61},
	pages={1-13},
	month={Mar.},
	}

@ARTICLE{9179756,
	author={Hong, Danfeng and Gao, Lianru and Hang, Renlong and Zhang, Bing and Chanussot, Jocelyn},
	journal={IEEE Geosci. Remote Sens. Lett.}, 
	title={Deep Encoder–Decoder Networks for Classification of Hyperspectral and LiDAR Data}, 
	year={2022},
	volume={19},
	number={},
	month={Aug.},
	pages={1-5},
	}

@ARTICLE{9598903,
	author={Wu, Xin and Hong, Danfeng and Chanussot, Jocelyn},
	journal={IEEE Trans. Geosci. Remote Sens.}, 
	title={Convolutional Neural Networks for Multimodal Remote Sensing Data Classification}, 
	year={2022},
	volume={60},
	number={},
	pages={1-10},
	month={Feb.},
	}

@article{tu2024ncglf2,
	title={NCGLF2: Network combining global and local features for fusion of multisource remote sensing data},
	author={Tu, Bing and Ren, Qi and Li, Jun and Cao, Zhaolou and Chen, Yunyun and Plaza, Antonio},
	journal={Inf. Fusion},
	volume={104},
	pages={102192},
	year={2024},
	month={Apr.},
	
}

@article{zhou2024tcpsnet,
	title={TCPSNet: Transformer and Cross-Pseudo-Siamese Learning Network for Classification of Multi-Source Remote Sensing Images},
	author={Zhou, Yongduo and Wang, Cheng and Zhang, Hebing and Wang, Hongtao and Xi, Xiaohuan and Yang, Zhou and Du, Meng},
	journal={Remote Sens.},
	volume={16},
	number={17},
	month = {Aug.},
	pages={3120},
	year={2024},

}

@article{li2023mixing,
	title={Mixing self-attention and convolution: A unified framework for multi-source remote sensing data classification},
	author={Li, Ke and Wang, Di and Wang, Xu and Liu, Gang and Wu, Zili and Wang, Quan},
	journal={IEEE Trans. Geosci. Remote Sens.},
	year={2023},
	volume={61},
	number={},
	pages={1-16},
	month={Sep.},

}

@ARTICLE{10438852,
	author={Ni, Kang and Wang, Duo and Zheng, Zhizhong and Wang, Peng},
	journal={IEEE J. Sel. Topics Appl. Earth Observ. Remote Sens.}, 
	title={MHST: Multiscale Head Selection Transformer for Hyperspectral and LiDAR Classification}, 
	year={2024},
	volume={17},
	number={},
	month={Feb.},
	pages={5470-5483},
}

@article{xie2024fusionmamba,
	title={Fusionmamba: Dynamic feature enhancement for multimodal image fusion with mamba},
	author={Xie, Xinyu and Cui, Yawen and Tan, Tao and Zheng, Xubin and Yu, Zitong},
	journal={Vis. Intell.},
	volume={2},
	number={1},
	pages={37},
	year={2024},
	publisher={Springer}
}

@ARTICLE{10738515,
	author={Zhang, Guanglian and Zhang, Zhanxu and Deng, Jiangwei and Bian, Lifeng and Yang, Chen},
	journal={IEEE Geosci. Remote Sens. Lett.}, 
	title={S2CrossMamba: Spatial–Spectral Cross-Mamba for Multimodal Remote Sensing Image Classification}, 
	year={2024},
	month={Oct.},
	volume={21},
	number={},
	pages={1-5},
	}

@ARTICLE{9755059,
	author={Xue, Zhixiang and Tan, Xiong and Yu, Xuchu and Liu, Bing and Yu, Anzhu and Zhang, Pengqiang},
	journal={IEEE Trans. Image Process.}, 
	title={Deep Hierarchical Vision Transformer for Hyperspectral and LiDAR Data Classification}, 
	year={2022},
	volume={31},
	number={},
	pages={3095-3110},
	month={Apr.},
	}

@inproceedings{xu2020learning,
	title={Learning in the frequency domain},
	author={Xu, Kai and Qin, Minghai and Sun, Fei and Wang, Yuhao and Chen, Yen-Kuang and Ren, Fengbo},
	pages={1740--1749},
	booktitle={Proc. IEEE Conf. Comput. Vis. Pattern Recognit. (CVPR)},
	year={2020},
}

@ARTICLE{1456290,
	author={Oppenheim, A.V. and Lim, J.S.},
	journal={Proc. IEEE}, 
	title={The importance of phase in signals}, 
	year={1981},
	volume={69},
	number={5},
	month={May},
	pages={529-541},
}

@inproceedings{yu2022frequency,
	title={Frequency and spatial dual guidance for image dehazing},
	author={Yu, Hu and Zheng, Naishan and Zhou, Man and Huang, Jie and Xiao, Zeyu and Zhao, Feng},
	booktitle={Eur. Conf. Comput. Vis},
	pages={181--198},
	year={2022},
}

@ARTICLE{9383794,
	author={Zhao, Xudong and Tao, Ran and Li, Wei and Philips, Wilfried and Liao, Wenzhi},
	journal={IEEE Trans. Geosci. Remote Sens.}, 
	title={Fractional Gabor Convolutional Network for Multisource Remote Sensing Data Classification}, 
	year={2022},
	volume={60},
	number={},
	month={Mar.},
	pages={1-18},
	}

@ARTICLE{8737724,
	author={Wu, Xin and Hong, Danfeng and Chanussot, Jocelyn and Xu, Yang and Tao, Ran and Wang, Yue},
	journal={IEEE Geosci. Remote Sens. Lett.}, 
	title={Fourier-Based Rotation-Invariant Feature Boosting: An Efficient Framework for Geospatial Object Detection}, 
	year={2020},
	volume={17},
	number={2},
	pages={302-306},
	month={Feb.},
	}

@inproceedings{zhao2022multisource,
	title={Multisource Cross-Scene Classification Using Fractional Fusion and Spatial-Spectral Domain Adaptation},
	author={Zhao, Xudong and Zhang, Mengmeng and Tao, Ran and Li, Wei and Liao, Wenzhi and Philips, Wilfried},
	booktitle={IEEE Geosci. Remote Sens. Symp.},
	pages={699--702},
	year={2022}
}

@inproceedings{zhao2022multisource2,
	title={Multisource remote sensing data classification using fractional Fourier transformer},
	author={Zhao, Xudong and Zhang, Mengmeng and Tao, Ran and Li, Wei and Liao, Wenzhi and Phlips, Wilfried},
	booktitle={IEEE Geosci. Remote Sens. Symp.},
	pages={823--826},
	year={2022},
	organization={IEEE}
}

@ARTICLE{9829269,
	author={Zhao, Xudong and Zhang, Mengmeng and Tao, Ran and Li, Wei and Liao, Wenzhi and Philips, Wilfried},
	journal={IEEE J. Sel. Topics Appl. Earth Observ. Remote Sens.},
	title={Cross-Domain Classification of Multisource Remote Sensing Data Using Fractional Fusion and Spatial-Spectral Domain Adaptation},
	year={2022},
	month={Jul.},
	volume={15},
	pages={5721-5733}
	}

@ARTICLE{4135672,
	author={Pattichis, Marios S. and Bovik, Alan C.},
	journal={IEEE Trans. Pattern Anal. Mach. Intell.}, 
	title={Analyzing Image Structure by Multidimensional Frequency Modulation}, 
	year={2007},
	volume={29},
	number={5},
	month={May},
	pages={753-766}
	}

@InProceedings{Mohla_2020_CVPR_Workshops,
	author = {Mohla, Satyam and Pande, Shivam and Banerjee, Biplab and Chaudhuri, Subhasis},
	title = {FusAtNet: Dual Attention Based SpectroSpatial Multimodal Fusion Network for Hyperspectral and LiDAR Classification},
	booktitle = {Proc. IEEE Conf. Comput. Vis. Pattern Recognit. Workshops (CVPRW)},
	pages={92--93},
	year = {2020}
}

@ARTICLE{9999457, 
	author={Zhao, Guangrui and Ye, Qiaolin and Sun, Le and Wu, Zebin and Pan, Chengsheng and Jeon, Byeungwoo}, 
	journal={IEEE Trans. Geosci. Remote Sens.}, 
	title={Joint Classification of Hyperspectral and LiDAR Data Using a Hierarchical CNN and Transformer}, year={2023}, 
	volume={61}, 
	number={}, 
	month={Jan.},
	pages={1-16} 
}

@ARTICLE{10314566,
	author={Lin, Junyan and Gao, Feng and Shi, Xiaochen and Dong, Junyu and Du, Qian},
	journal={IEEE Trans. Geosci. Remote Sens.}, 
	title={SS-MAE: Spatial–Spectral Masked Autoencoder for Multisource Remote Sensing Image Classification}, 
	year={2023},
	volume={61},
	number={},
	pages={1-14},
	month={Nov.}
	}

@ARTICLE{10167673,
	author={Wang, Junjie and Li, Wei and Wang, Yinjian and Tao, Ran and Du, Qian},
	journal={IEEE Trans. Neural Netw. Learn. Syst.},
	title={Representation-Enhanced Status Replay Network for Multisource Remote-Sensing Image Classification},
	year={2023},
	month={Jun.},
	pages={1-13},
}

@ARTICLE{11386975,
	author={Ren, Bo and Wang, Qianfang and Liu, Bo and Hou, Biao and Yang, Chen and Jiao, Licheng},
	journal={IEEE Trans. Geosci. Remote Sens.}, 
	title={WHFNet: A Wavelet-Driven Heterogeneous Fusion Network for High-Frequency Enhanced Optical–SAR Remote Sensing Segmentation}, 
	year={2026},
	volume={64},
	number={},
	pages={1-17},
	}

@ARTICLE{10945964,
	author={Chen, Yaxiong and Wang, Qicong and Zhao, Yichen and Xiong, Shengwu and Lu, Xiaoqiang},
	journal={IEEE Trans. Geosci. Remote Sens.}, 
	title={Bilinear Parallel Fourier Transformer for Multimodal Remote Sensing Classification}, 
	year={2025},
	volume={63},
	number={},
	pages={1-14},
	}

@inproceedings{yang2025ffr,
	title={Ffr: Frequency feature rectification for weakly supervised semantic segmentation},
	author={Yang, Ziqian and Zhao, Xinqiao and Wang, Xiaolei and Zhang, Quan and Xiao, Jimin},
	booktitle={Proc. IEEE Conf. Comput. Vis. Pattern Recognit. (CVPR)},
	pages={30261--30270},
	year={2025}
}

@INPROCEEDINGS{10943595,
	author={Patro, Badri N. and Namboodiri, Vinay P. and Agneeswaran, Vijay S.},
	booktitle={Proc. IEEE/CVF Winter Conf. Appl. Comput. Vis. (WACV)}, 
	title={SpectFormer: Frequency and Attention is what you need in a Vision Transformer}, 
	year={2025},
	volume={},
	number={},
	pages={9543-9554},
	}

@inproceedings{chen2026f2net,
	title={F2Net: A frequency-fused network for ultra-high resolution remote sensing segmentation},
	author={Chen, Hengzhi and Feng, Liqian and Wu, Wenhua and Zhu, Xiaogang and Wu, Qiuxia and Shan, Lianlei and Hu, Kun},
	booktitle={Proc. IEEE Conf. Comput. Vis. Pattern Recognit. (CVPR)},
	pages={13275--13284},
	year={2026}
}
	
\end{document}